\documentclass[10pt,twocolumn,letterpaper]{article}

\usepackage{authblk}
\usepackage{wacv}  
\usepackage{calrsfs}

\usepackage{graphicx}
\usepackage{amsmath}
\usepackage{amssymb}
\usepackage{booktabs}
\usepackage{float}
\usepackage{multirow}
\usepackage[pagebackref,breaklinks,colorlinks]{hyperref}

\usepackage[capitalize]{cleveref}
\crefname{section}{Sec.}{Secs.}
\Crefname{section}{Section}{Sections}
\Crefname{table}{Table}{Tables}
\crefname{table}{Tab.}{Tabs.}

\def\wacvPaperID{1881} 
\def\confName{WACV}
\def\confYear{2024}

\usepackage{graphicx}
\usepackage{caption}
\usepackage{subcaption}
\usepackage{tikz}
\usepackage{pgfplots}
\usetikzlibrary{spy,calc}
\usepackage{hyperref}

\pgfplotsset{compat=1.18}
\newif\ifblackandwhitecycle
\gdef\patternnumber{0}

\pgfkeys{/tikz/.cd,
    zoombox paths/.style={
        draw=orange,
        very thick
    },
    black and white/.is choice,
    black and white/.default=static,
    black and white/static/.style={ 
        draw=white,   
        zoombox paths/.append style={
            draw=white,
            postaction={
                draw=black,
                loosely dashed
            }
        }
    },
    black and white/static/.code={
        \gdef\patternnumber{1}
    },
    black and white/cycle/.code={
        \blackandwhitecycletrue
        \gdef\patternnumber{1}
    },
    black and white pattern/.is choice,
    black and white pattern/0/.style={},
    black and white pattern/1/.style={    
            draw=white,
            postaction={
                draw=black,
                dash pattern=on 2pt off 2pt
            }
    },
    black and white pattern/2/.style={    
            draw=white,
            postaction={
                draw=black,
                dash pattern=on 4pt off 4pt
            }
    },
    black and white pattern/3/.style={    
            draw=white,
            postaction={
                draw=black,
                dash pattern=on 4pt off 4pt on 1pt off 4pt
            }
    },
    black and white pattern/4/.style={    
            draw=white,
            postaction={
                draw=black,
                dash pattern=on 4pt off 2pt on 2 pt off 2pt on 2 pt off 2pt
            }
    },
    zoomboxarray inner gap/.initial=5pt,
    zoomboxarray columns/.initial=2,
    zoomboxarray rows/.initial=2,
    subfigurename/.initial={},
    figurename/.initial={zoombox},
    zoomboxarray/.style={
        execute at begin picture={
            \begin{scope}[
                spy using outlines={%
                    zoombox paths,
                    width=\imagewidth / \pgfkeysvalueof{/tikz/zoomboxarray columns} - (\pgfkeysvalueof{/tikz/zoomboxarray columns} - 1) / \pgfkeysvalueof{/tikz/zoomboxarray columns} * \pgfkeysvalueof{/tikz/zoomboxarray inner gap} -\pgflinewidth,
                    height=\imageheight / \pgfkeysvalueof{/tikz/zoomboxarray rows} - (\pgfkeysvalueof{/tikz/zoomboxarray rows} - 1) / \pgfkeysvalueof{/tikz/zoomboxarray rows} * \pgfkeysvalueof{/tikz/zoomboxarray inner gap}-\pgflinewidth,
                    magnification=3,
                    every spy on node/.style={
                        zoombox paths
                    },
                    every spy in node/.style={
                        zoombox paths
                    }
                }
            ]
        },
        execute at end picture={
            \end{scope}
            \node at (image.north) [anchor=north,inner sep=0pt] {\subcaptionbox{\label{\pgfkeysvalueof{/tikz/figurename}-image}}{\phantomimage}};
            \node at (zoomboxes container.north) [anchor=north,inner sep=0pt] {\subcaptionbox{\label{\pgfkeysvalueof{/tikz/figurename}-zoom}}{\phantomimage}};
     \gdef\patternnumber{0}
        },
        spymargin/.initial=0.5em,
        zoomboxes xshift/.initial=1,
        zoomboxes right/.code=\pgfkeys{/tikz/zoomboxes xshift=1},
        zoomboxes left/.code=\pgfkeys{/tikz/zoomboxes xshift=-1},
        zoomboxes yshift/.initial=0,
        zoomboxes above/.code={
            \pgfkeys{/tikz/zoomboxes yshift=1},
            \pgfkeys{/tikz/zoomboxes xshift=0}
        },
        zoomboxes below/.code={
            \pgfkeys{/tikz/zoomboxes yshift=-1},
            \pgfkeys{/tikz/zoomboxes xshift=0}
        },
        caption margin/.initial=4ex,
    },
    adjust caption spacing/.code={},
    image container/.style={
        inner sep=0pt,
        at=(image.north),
        anchor=north,
        adjust caption spacing
    },
    zoomboxes container/.style={
        inner sep=0pt,
        at=(image.north),
        anchor=north,
        name=zoomboxes container,
        xshift=\pgfkeysvalueof{/tikz/zoomboxes xshift}*(\imagewidth+\pgfkeysvalueof{/tikz/spymargin}),
        yshift=\pgfkeysvalueof{/tikz/zoomboxes yshift}*(\imageheight+\pgfkeysvalueof{/tikz/spymargin}+\pgfkeysvalueof{/tikz/caption margin}),
        adjust caption spacing
    },
    calculate dimensions/.code={
        \pgfpointdiff{\pgfpointanchor{image}{south west} }{\pgfpointanchor{image}{north east} }
        \pgfgetlastxy{\imagewidth}{\imageheight}
        \global\let\imagewidth=\imagewidth
        \global\let\imageheight=\imageheight
        \gdef\columncount{1}
        \gdef\rowcount{1}
        
    },
    image node/.style={
        inner sep=0pt,
        name=image,
        anchor=south west,
        append after command={
            [calculate dimensions]
            node [image container,subfigurename=\pgfkeysvalueof{/tikz/figurename}-image] {\phantomimage}
            node [zoomboxes container,subfigurename=\pgfkeysvalueof{/tikz/figurename}-zoom] {\phantomimage}
        }
    },
    color code/.style={
        zoombox paths/.append style={draw=#1}
    },
    connect zoomboxes/.style={
    spy connection path={\draw[draw=none,zoombox paths] (tikzspyonnode) -- (tikzspyinnode);}
    },
    help grid code/.code={
        \begin{scope}[
                x={(image.south east)},
                y={(image.north west)},
                font=\footnotesize,
                help lines,
                overlay
            ]
            \foreach \x in {0,1,...,9} { 
                \draw(\x/10,0) -- (\x/10,1);
                \node [anchor=north] at (\x/10,0) {0.\x};
            }
            \foreach \y in {0,1,...,9} {
                \draw(0,\y/10) -- (1,\y/10);                        \node [anchor=east] at (0,\y/10) {0.\y};
            }
        \end{scope}    
    },
    help grid/.style={
        append after command={
            [help grid code]
        }
    },
}

\newcommand\phantomimage{%
    \phantom{%
        \rule{\imagewidth}{\imageheight}%
    }%
}
\newcommand\zoombox[2][]{
    \begin{scope}[zoombox paths]
        \pgfmathsetmacro\xpos{
            (\columncount-1)*(\imagewidth / \pgfkeysvalueof{/tikz/zoomboxarray columns} + \pgfkeysvalueof{/tikz/zoomboxarray inner gap} / \pgfkeysvalueof{/tikz/zoomboxarray columns} ) + \pgflinewidth
        }
        \pgfmathsetmacro\ypos{
            (\rowcount-1)*( \imageheight / \pgfkeysvalueof{/tikz/zoomboxarray rows} + \pgfkeysvalueof{/tikz/zoomboxarray inner gap} / \pgfkeysvalueof{/tikz/zoomboxarray rows} ) + 0.5*\pgflinewidth
        }
        \edef\dospy{\noexpand\spy [
            #1,
            zoombox paths/.append style={
                black and white pattern=\patternnumber
            },
            every spy on node/.append style={#1},
            x=\imagewidth,
            y=\imageheight
        ] on (#2) in node [anchor=north west] at ($(zoomboxes container.north west)+(\xpos pt,-\ypos pt)$);}
        \dospy
        \pgfmathtruncatemacro\pgfmathresult{ifthenelse(\columncount==\pgfkeysvalueof{/tikz/zoomboxarray columns},\rowcount+1,\rowcount)}
        \global\let\rowcount=\pgfmathresult
        \pgfmathtruncatemacro\pgfmathresult{ifthenelse(\columncount==\pgfkeysvalueof{/tikz/zoomboxarray columns},1,\columncount+1)}
        \global\let\columncount=\pgfmathresult
        \ifblackandwhitecycle
            \pgfmathtruncatemacro{\newpatternnumber}{\patternnumber+1}
            \global\edef\patternnumber{\newpatternnumber}
        \fi
    \end{scope}
}

\pgfkeys{/tikz/.cd,
    zoombox 2 paths/.style={
        draw=orange,
        very thick
    },
    black and white/.is choice,
    black and white/.default=static,
    black and white/static/.style={ 
        draw=white,   
        zoombox 2 paths/.append style={
            draw=white,
            postaction={
                draw=black,
                loosely dashed
            }
        }
    },
    black and white/static/.code={
        \gdef\patternnumber{1}
    },
    black and white/cycle/.code={
        \blackandwhitecycletrue
        \gdef\patternnumber{1}
    },
    black and white pattern/.is choice,
    black and white pattern/0/.style={},
    black and white pattern/1/.style={    
            draw=white,
            postaction={
                draw=black,
                dash pattern=on 2pt off 2pt
            }
    },
    black and white pattern/2/.style={    
            draw=white,
            postaction={
                draw=black,
                dash pattern=on 4pt off 4pt
            }
    },
    black and white pattern/3/.style={    
            draw=white,
            postaction={
                draw=black,
                dash pattern=on 4pt off 4pt on 1pt off 4pt
            }
    },
    black and white pattern/4/.style={    
            draw=white,
            postaction={
                draw=black,
                dash pattern=on 4pt off 2pt on 2 pt off 2pt on 2 pt off 2pt
            }
    },
    zoomboxarray 2 inner gap/.initial=5pt,
    zoomboxarray 2 columns/.initial=3,
    zoomboxarray 2 rows/.initial=2,
    subfigurename/.initial={},
    figurename/.initial={zoomboxtwo},
    zoomboxarray 2/.style={
        execute at begin picture={
            \begin{scope}[
                spy using outlines={%
                    zoombox 2 paths,
                    width=\imagewidth / 1 / \pgfkeysvalueof{/tikz/zoomboxarray 2 columns} - (\pgfkeysvalueof{/tikz/zoomboxarray 2 columns} - 1) / \pgfkeysvalueof{/tikz/zoomboxarray 2 columns} * \pgfkeysvalueof{/tikz/zoomboxarray 2 inner gap} -\pgflinewidth,
                    height=\imageheight / 1.5 / \pgfkeysvalueof{/tikz/zoomboxarray 2 rows} - (\pgfkeysvalueof{/tikz/zoomboxarray 2 rows} - 1) / \pgfkeysvalueof{/tikz/zoomboxarray 2 rows} * \pgfkeysvalueof{/tikz/zoomboxarray 2 inner gap}-\pgflinewidth,
                    magnification=3,
                    every spy on node/.style={
                        zoombox 2 paths
                    },
                    every spy in node/.style={
                        zoombox 2 paths
                    }
                }
            ]
        },
        execute at end picture={
            \end{scope}
            \node at (image 2.north) [anchor=north,inner sep=0pt] {\subcaptionbox{\label{\pgfkeysvalueof{/tikz/figurename}-image 2}}{\phantomimage}};
            \node at (zoomboxes 2 container.north) [anchor=north,inner sep=0pt] {\subcaptionbox{\label{\pgfkeysvalueof{/tikz/figurename}-zoom 2}}{\phantomimagetwo}};
     \gdef\patternnumber{0}
        },
        spymargin/.initial=0.5em,
        zoomboxes 2 xshift/.initial=1,
        zoomboxes 2 right/.code=\pgfkeys{/tikz/zoomboxes 2 xshift=1},
        zoomboxes 2 left/.code=\pgfkeys{/tikz/zoomboxes 2 xshift=-1},
        zoomboxes 2 yshift/.initial=0,
        zoomboxes 2 above/.code={
            \pgfkeys{/tikz/zoomboxes 2 yshift=1},
            \pgfkeys{/tikz/zoomboxes 2 xshift=0}
        },
        zoomboxes 2 below/.code={
            \pgfkeys{/tikz/zoomboxes 2 yshift=-1},
            \pgfkeys{/tikz/zoomboxes 2 xshift=0}
        },
        caption margin/.initial=4ex,
    },
    adjust caption spacing/.code={},
    image container 2/.style={
        inner sep=0pt,
        at=(image 2.north),
        anchor=north,
        adjust caption spacing
    },
    zoomboxes 2 container/.style={
        inner sep=0pt,
        at=(image 2.north),
        anchor=north,
        name=zoomboxes 2 container,
        xshift=\pgfkeysvalueof{/tikz/zoomboxes 2 xshift}*(\imagewidth+\pgfkeysvalueof{/tikz/spymargin}),
        yshift=\pgfkeysvalueof{/tikz/zoomboxes 2 yshift}*(\imageheight+\pgfkeysvalueof{/tikz/spymargin}+\pgfkeysvalueof{/tikz/caption margin}),
        adjust caption spacing
    },
    calculate dimensions 2/.code={
        \pgfpointdiff{\pgfpointanchor{image 2}{south west} }{\pgfpointanchor{image 2}{north east} }
        \pgfgetlastxy{\imagewidth}{\imageheight}
        \global\let\imagewidth=\imagewidth
        \global\let\imageheight=\imageheight
        \gdef\columncount{1}
        \gdef\rowcount{1}
        
    },
    image node 2/.style={
        inner sep=0pt,
        name=image 2,
        anchor=south west,
        append after command={
            [calculate dimensions 2]
            node [image container 2,subfigurename=\pgfkeysvalueof{/tikz/figurename}-image 2] {\phantomimage}
            node [zoomboxes 2 container,subfigurename=\pgfkeysvalueof{/tikz/figurename}-zoom] {\phantomimagetwo}
        }
    },
    color code 2/.style={
        zoombox 2 paths/.append style={draw=#1}
    },
    connect zoomboxes 2/.style={
    spy connection path={\draw[draw=none,zoombox 2 paths] (tikzspyonnode) -- (tikzspyinnode);}
    },
    help grid code/.code={
        \begin{scope}[
                x={(image 2.south east)},
                y={(image 2.north west)},
                font=\footnotesize,
                help lines,
                overlay
            ]
            \foreach \x in {0,1,...,9} { 
                \draw(\x/10,0) -- (\x/10,1);
                \node [anchor=north] at (\x/10,0) {0.\x};
            }
            \foreach \y in {0,1,...,9} {
                \draw(0,\y/10) -- (1,\y/10);                        \node [anchor=east] at (0,\y/10) {0.\y};
            }
        \end{scope}    
    },
    help grid/.style={
        append after command={
            [help grid code]
        }
    },
}
\newcommand\phantomimagetwo{%
    \phantom{%
        \rule{8.25cm}{6.85cm}%
    }%
}
\newcommand\zoomboxtwo[2][]{
    \begin{scope}[zoombox 2 paths]
        \pgfmathsetmacro\xpos{
            (\columncount-1)*(\imagewidth / 1 / \pgfkeysvalueof{/tikz/zoomboxarray 2 columns} + \pgfkeysvalueof{/tikz/zoomboxarray 2 inner gap} / \pgfkeysvalueof{/tikz/zoomboxarray 2 columns} ) + \pgflinewidth
        }
        \pgfmathsetmacro\ypos{
            (\rowcount-1)*( \imageheight / 1.5 / \pgfkeysvalueof{/tikz/zoomboxarray 2 rows} + \pgfkeysvalueof{/tikz/zoomboxarray 2 inner gap} / \pgfkeysvalueof{/tikz/zoomboxarray 2 rows} ) + 0.5*\pgflinewidth
        }
        \edef\dospy{\noexpand\spy [
            #1,
            zoombox 2 paths/.append style={
                black and white pattern=\patternnumber
            },
            every spy on node/.append style={#1},
            x=\imagewidth,
            y=\imageheight
        ] on (#2) in node [anchor=north west] at ($(zoomboxes 2 container.north west)+(\xpos pt,-\ypos pt)$);}
        \dospy
        \pgfmathtruncatemacro\pgfmathresult{ifthenelse(\columncount==\pgfkeysvalueof{/tikz/zoomboxarray 2 columns},\rowcount+1,\rowcount)}
        \global\let\rowcount=\pgfmathresult
        \pgfmathtruncatemacro\pgfmathresult{ifthenelse(\columncount==\pgfkeysvalueof{/tikz/zoomboxarray 2 columns},1,\columncount+1)}
        \global\let\columncount=\pgfmathresult
        \ifblackandwhitecycle
            \pgfmathtruncatemacro{\newpatternnumber}{\patternnumber+1}
            \global\edef\patternnumber{\newpatternnumber}
        \fi
    \end{scope}
}

\begin{document}

\title{Face Cartoonisation For Various Poses Using StyleGAN}
\author[1]{Kushal Jain}
\author[1]{Ankith Varun J}
\author[1]{Anoop Namboodiri}
\affil[1]{CVIT, IIIT-Hyderabad}


\twocolumn[{
\maketitle
\begin{center}
\centering
\captionsetup{type=figure}\addtocounter{figure}{-1}
\begin{tikzpicture}[zoomboxarray]
    \node [image node] { \includegraphics[width=0.49\textwidth]{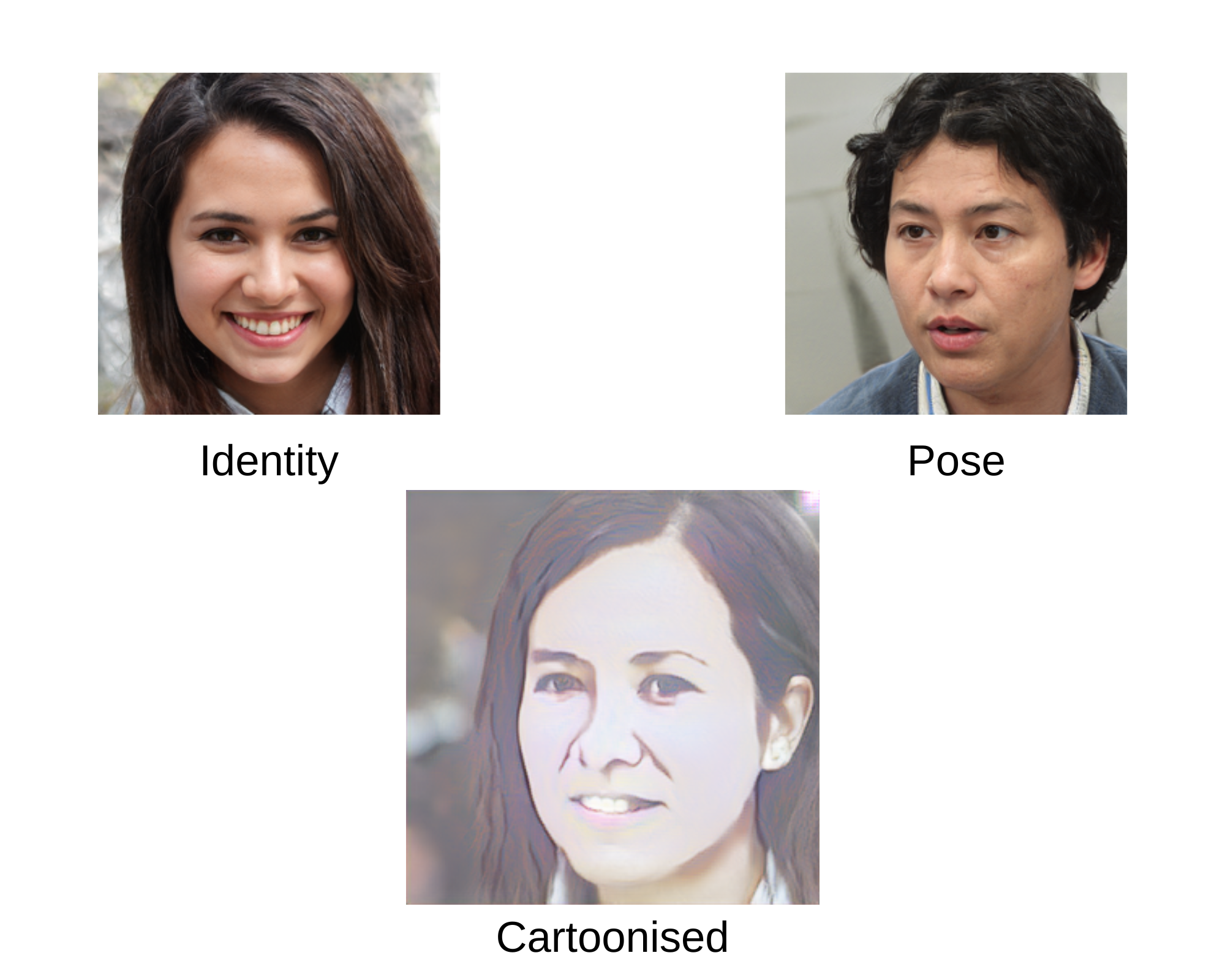} };
    \zoombox[magnification=6,color code=red]{0.535,0.45}
    \zoombox[magnification=8,color code=blue]{0.63,0.29}
    \zoombox[magnification=8,color code=green]{0.495,0.25}
    \zoombox[magnification=9,color code=yellow]{0.43,0.14}
\end{tikzpicture}
\captionof{figure}{A simple illustration of our method. (a) shows the input identity and pose images (top) along with our cartoonised output (bottom) (b) clear edges, smooth and flat textures.}
\label{illus}
\end{center}
}]

\begin{abstract}
   This paper presents an innovative approach to achieve face cartoonisation while preserving the original identity and accommodating various poses. Unlike previous methods in this field that relied on conditional-GANs, which posed challenges related to dataset requirements and pose training, our approach leverages the expressive latent space of StyleGAN. We achieve this by introducing an encoder that captures both pose and identity information from images and generates a corresponding embedding within the StyleGAN latent space. By subsequently passing this embedding through a pre-trained generator, we obtain the desired cartoonised output.

While many other approaches based on StyleGAN necessitate a dedicated and fine-tuned StyleGAN model, our method stands out by utilizing an already-trained StyleGAN designed to produce realistic facial images. We show by extensive experimentation how our encoder adapts the StyleGAN output to better preserve identity when the objective is cartoonisation.
   Our code will be released upon acceptance.\\
\end{abstract} 

\section{Introduction}
\label{sec:intro}
Generative Adversarial Networks (GANs) \cite{goodfellow2014generative} have demonstrated remarkable success in a wide range of computer vision applications, including cartoonisation \cite{Wang_2020_CVPR, cartoonGAN}, video generation, image-to-image translation \cite{UNIT,pix2pix2017,cyclegan}. Over the past few years, there has been a rapid improvement in the quality of images synthesized by GANs. From the seminal DCGAN framework \cite{Radford2015UnsupervisedRL} in 2015, the current state-of-the-art GANs can generate images at much higher resolutions and produce significantly more realistic results. A notable advancement in this trajectory is StyleGAN \cite{Karras_2019_CVPR}, which introduces an intermediate latent space denoted as $W$. This latent space constitutes a pivotal innovation that holds immense potential for enabling controlled image modifications within unsupervised settings. This intermediate space, makes it feasible to conduct fine-grained alterations to images, allowing for an unprecedented level of manipulation while maintaining high fidelity.\\ 
Our method utilizes the rich latent space ($W$) of a StyleGAN pretrained on FFHQ \cite{Karras_2019_CVPR} dataset paired with a GAN-based cartoon generator \cite{Wang_2020_CVPR} to generate cartoonized faces. These both combined make our generator setup. In contrast to many other artistic styles that involve adding smaller intricate textures such as brush strokes or shading lines, cartoon images exhibit a distinct feel that is achieved by simplifying and abstracting elements from real-world photographs. Common features of cartoon images, also pointed out by authors in \cite{cartoonGAN}, are well-defined edges, a consistent application of smooth color shading, and relatively plain textures. These features sets cartoon images apart from various other forms of artwork that may prioritize different techniques and aesthetics.  \\
We found that by integrating the outputs of the cartoon generator to calculate loss, we imbue our encoder with a sense of what the desired cartoonised version of the input image should entail. This supplemental supervision serves as an anchor during the loss computation, enabling our encoder to send a more cartoonised signal to StyleGAN. Furtheremore, it affords us the ability to sidestep the otherwise demanding process of fine-tuning the StyleGAN model, which requires considerable computational resources. By leveraging the features learnt by CartoonGAN framework during training with unpaired datasets , we not only streamline the training process for our generator setup but our results also show strong cartoon characteristics (see Fig. \ref{illus}). \\
Our method takes the  identity image ($I_{id}$) and the pose image ($I_{p}$) as inputs and extracts the required features from those images using encoders namely $E_{id}$ for identity and $E_{p}$ for pose. These features are passed through a Multi-Layer Perceptron which is trained to generate a vector $w \in W$ which corresponds to the cartoonised version of $I_{id}$ in the pose of $I_{p}$. When we pass this vector through our generator setup to get the output cartoon face. The main contributions that we making in this paper are as follows : 
\begin{enumerate}
    \item We propose a method to achieve cartoonisation by learning the seperate representations for identity and pose in the $W$ latent space of StyleGAN allowing us to generate various poses and identities without class-supervision.
    \item We use a readily available StyleGAN pretrained on FFHQ dataset with a cartoon generator which allows for skipping the time-consuming fine-tuning process.
    \item We show that the distribution our encoder learns in the latent space, is better for encoding identity when cartoonisation is the objective. 
\end{enumerate}

\section{Related Works}
\label{sec:related}
\subsection{Conditional-GANs}
 In cGANs \cite{Mirza2014ConditionalGA}, the generator network takes both random noise and conditional information as input to produce contextually relevant outputs. Isola et al. \cite{pix2pix2017} extended cGANs to general image to image translation but it required paired datasets. Recently many works have successfully achieved image-to-image translation with unpaired datasets using cycle consistency loss \cite{cyclegan} and other losses \cite{UGATIT, MUNIT, DualGAN}. The utilization of such image-to-image translation techniques finds relevance  in scenarios involving inpainting, where the transformation occurs between unconventional domains, like the shift from a 3D-2D rendered domain to a photo-realistic domain \cite{RotateAndRender}.

In the domain of face manipulation, there exists a diverse array of techniques classified based on whether they prioritize preserving identity \cite{FaceIDGANLA} or manipulating other facial attributes \cite{UNIT}. Several works employ conditional GANs to translate between discrete domains such as age or emotional expression, necessitating labeled datasets but capable of operating on unseen images \cite{FUNIT,UGATIT,cartoonGAN}.  \\
Unlike existing approaches that rely on black-box models and loss terms for network training, our proposed method takes a different approach by learning disentangled representations in the StyleGAN latent space\cite{Richardson2020EncodingIS}. This helps in enforcing the network to learn distinct features with separate objectives, making the learning process more controllable and tunable.

\subsection{Latent Space of StyleGAN}
The exploration and control of latent spaces in GANs have been a subject of significant research. One of the tasks that has started to garner attention is GAN inversion which aims to find the latent vector in a pretrained GAN, that accurately reconstructs a given image \cite{inversion}. GAN inversion has been widely explored using StyleGAN \cite{Karras_2019_CVPR}, because of its rich and expressive latent space.\\
Recently, GAN inversion has been achieved by mainly two methods, by directly optimizing the latent vector to minimize reconstruction error for every image \cite{Abdal_2019_ICCV,inversion_new} or by training an encoder to map images to the latent space \cite{Richardson2020EncodingIS}. We draw inspiration from \cite{Richardson2020EncodingIS} and make an encoder that directly maps the input conditions onto the StyleGAN latent space, doing away from ``invert first edit later" paradigms \cite{EditingIS}. \\
Through explorations of the latent space, several works have effectively showcased the latent vectors' capacity to facilitate seamless transitions among distinct facial characteristics, expressions, and poses \cite{pix2style2pix, ostec, StyleFusion}. This phenomenon highlights the exceptional ability of StyleGAN's latent space to disentangle and represent semantic attributes in a coherent manner.
\begin{figure*}[ht]
     \centering
    \includegraphics[width=0.9\textwidth]{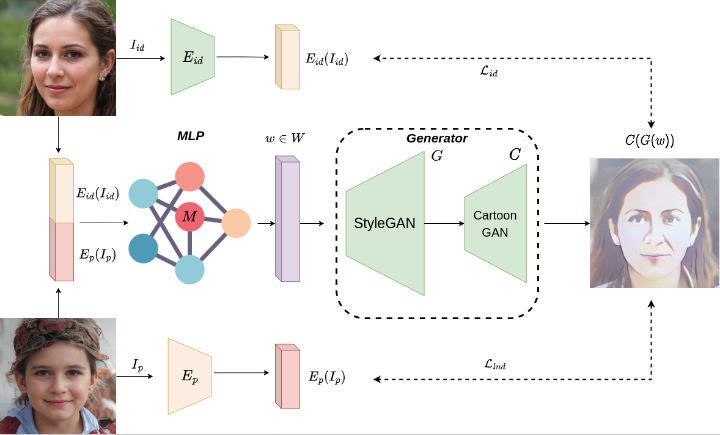}
    \caption{Our model architecture. All the models in green are pretrained, while the pose encoder in yellow is being fine-tuned and the MLP is trained from scratch. Data flow is marked with solid lines and losses are marked with dashed lines. Bold-dashed lines enclose the generator setup. Input images $I_{id}$ and $I_{p}$ are encoded using $E_{id}$ and $E_{p}$ respectively, to give embeddings. The embeddings are concatenated and passed through the MLP which maps it to the $W$ latent space of StyleGAN. The $w$ vector generated is passed through our generator setup to give the final output $G(C(w))$. $\mathcal{L}_{id}$ ensures that identity is preserved, $\mathcal{L}_{lnd}$ enforces pose translation. Another loss term $\mathcal{L}_{rec}$ is also used but is not shown here.}
    \label{modelarch}
\end{figure*}
\section{Method}
\label{sec:method}
Given two images $I_d$ and $I_p$, our method generates a cartoon image representing the identity of $I_d$ and other attributes, primarily pose and expression of $I_p$. As show in Fig. \ref{modelarch}, our model roughly comprises of encoders, a Multi-Layer Perceptron (MLP) and generators. On a high level, the encoders capture the respective features of the input images which are then mapped to the latent space of the generators by the MLP, finally resulting in the required cartoon image.

The input images ($I_{id}, I_{p} $) serve as the foundation from which we derive the necessary information to guide our process. To distill the salient characteristics from these images, we employ dedicated encoders. These encoders are adept at extracting pertinent features from their respective input images, hence their embeddings encapsulate the intrinsic elements that contribute to identity and pose attributes.

 Essentially, the MLP functions as an interpreter, translating the extracted features into a concise and expressive instruction set for our subsequent steps.
 We adopt a strategy where the generator, depicted in green (Fig. \ref{modelarch}), is fully pretrained, while only the MLP remains untrained. Simultaneously, we focus on refining the performance of the pose encoder, denoted as $E_p$ and highlighted in yellow. This involves finetuning of the encoder's final layers to enhance its capability to capture pose information accurately. The overarching goal of this finetuning endeavor is to achieve a more effective and nuanced translation of pose-related data, contributing to the overall success of our method, also observed by \cite{Nitzan2020D}. 
\subsection{Encoders}

In our approach, we integrate identity and pose information using well-established models. For identity encoding, we employ a pre-trained Arcface face recognition model \cite{ArcFaceAA}, extracting activations from its penultimate layer to create an identity embedding vector ($E_{id}(I_{id})$).\\   
To encode pose we use the pretrained VGG \cite{VGG}, trained on face images, but the model is also being finetuned to serve our purpose better. We use the penultimate layer activations as our pose embedding vector ($E_{p}(I_{p})$). We concatenate the two embeddings and pass it through the MLP to get the $w$ vector.
$$ w = M([E_{p}(I_{p}) , E_{id}(I_{id}) ])$$
We also make use of a landmark encoder $E_{lnd}$ in the training process in order to captures facial landmarks of realistic images.
\subsection{Generator}
Our generator comprises of two essential components. Firstly, we employ a StyleGAN ($G$) that has been pre-trained on the FFHQ dataset \cite{Karras_2019_CVPR}. This acts like a pre-cartoonisation backbone. Our aim here is not to generate realistic faces but faces that cartoonise well in the subsequent step. \\
Now to transfer the style of our generated image from real to cartoon, we pass our image through a GAN-based cartoon generator called White Box Cartooniser ($C$), as detailed in the work by Wang and colleagues \cite{Wang_2020_CVPR}. Their work builds on the foundational work of CartoonGAN \cite{cartoonGAN} that introduced a novel edge loss to cartoonise real-life images. White Box Cartooniser on the other hand breaks down the problem into surface representation, structure representation and texture representation and optimises for them individually which makes the framework more controllable and tunable.   \\
These image to image translation models have learnt many crucial features needed for our task. Thus, the decision to leverage these models aligns intuitively with the requirements of our task. We pass the $w$ vector through the $G$ and then through $C$ to get our output image.
$$I_{out} = C(G(w))$$
\subsection{Losses}
\label{sec:losses}

\textbf{Identity Loss : } The process of transforming facial images into their cartoonised counterparts involves a delicate balance between creative distortion and maintaining the essential identity of the subject. To achieve this delicate equilibrium, we introduce a pivotal identity loss term, formulated as an $L_{1}$ cycle consistency loss between $C(I_{id})$ and $I_{out}$ as follows. 
\begin{equation}
\mathcal{L}_{id} = \lVert E_{id}(C(I_{id})) - E_{id}(I_{out}) \rVert _{1}  
\end{equation}
 Cartoonisation involves a deliberate departure from photo-realism, introducing stylistic exaggerations and distortions to achieve the desired artistic effect, which may degrade identity preservation. Therefore, we align our approach with the objective of preserving the essential identity attributes in the transformed cartoon space.
\begin{figure}[H]
     \centering
    \includegraphics[width=\columnwidth]{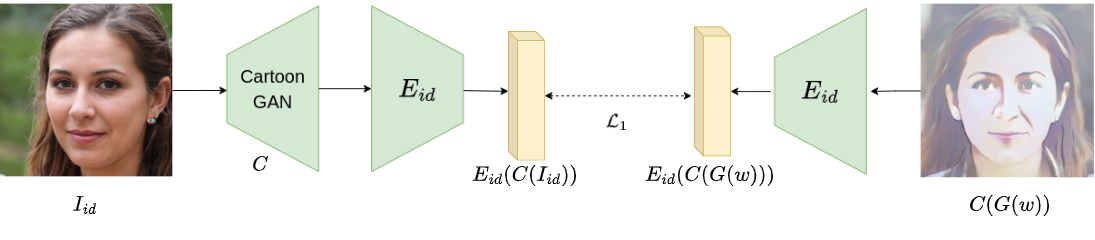}
    \caption{A visualisation of identity loss calculation. $E_{id}$ captures the identity features from the generated image and the cartoonised version of the identity image to compute the loss.}
    \label{idloss}
\end{figure}

\textbf{Landmark Loss : }The human visual system excels at discerning identity-defining details like facial arrangement and subtle variations, hence a landmarks loss becomes crucial. The importance of this loss can also be seen in Sec. \ref{abl}, where we show that by not using landmark loss the MLP is unable to transfer pose accurately. As a result, we incorporate a sparse $L_{2}$ cycle consistency landmark loss as given below. 
\begin{equation}
\mathcal{L}_{lnd} = \lVert E_{lnd}(C(I_{p})) - E_{lnd}(I_{out}) \rVert _{2}  
\end{equation}

\begin{figure}[H]
     \centering
    \includegraphics[width=\columnwidth]{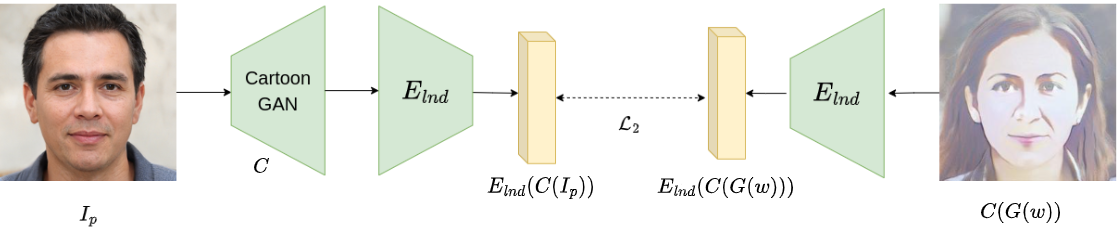}
    \caption{A visualisation of landmark loss computation. $E_{lnd}$ encodes the facial landmarks in the cartoon space to ensure consistency and penalise differences in pose.}
    \label{lndloss}
\end{figure}

\textbf{Reconstruction Loss : }
In addition to identity and pose, preservation of other factors such as illumination and colour, is instrumental in generating a good generalized cartoon image especially in the scenario when $I_{id}$ and $I_{p}$ are identical. In our proposed architecture, whenever $I_{id} = I_{p}$, the MLP is tasked with inversion so using a pixel-level reconstruction loss makes intuitive sense. Motivated by the results of \cite{Nitzan2020D} we use a \textit{mix} loss $\mathcal{L}_{mix}$ \cite{ssim} to encourage a more refined cartoonisation preserving non-facial attributes.  $\mathcal{L}_{mix}$ is a weighted sum of $L_1$ loss and MS-SSIM loss with $\alpha$ being a hyper-parameter.
\begin{equation}
\mathcal{L}_{mix} = \alpha(1 - MS{\text -} SSIM(I_{p},I_{id})) + (1-\alpha)\lVert I_{p} - I_{id} \rVert_{1}
\end{equation}
We impose a constraint on the application of this loss, limiting it exclusively to cases where the identity and pose images are identical. This deliberate constraint is designed to prevent the reconstruction of identity features from $I_p$ within the resulting cartoon image. The below redefinition of $\mathcal{L}_{mix}$ takes this into account.
\begin{equation}
 \mathcal{L}_{rec} =
    \begin{cases}
     \mathcal{L}_{mix}  & \text{if $I_{id} = I_{p}$}\\
      0 & \text{otherwise}\\
    \end{cases}       
\end{equation}
\begin{figure*}[ht]
     \centering
    \includegraphics[width=\textwidth]{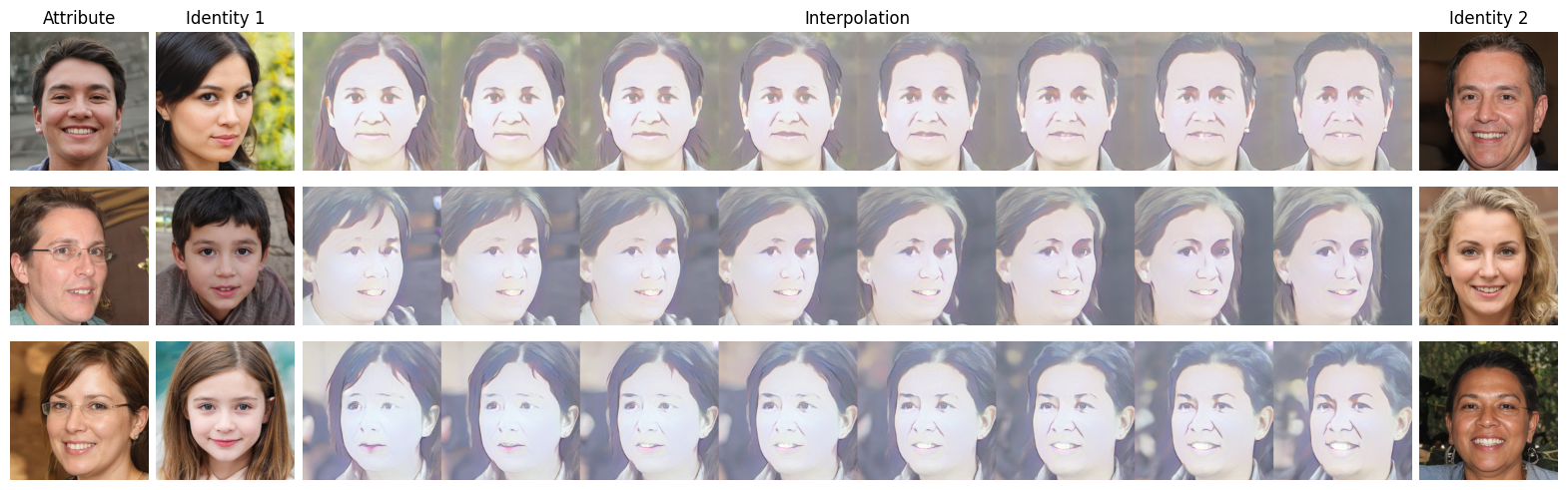}
    \includegraphics[width=\textwidth]{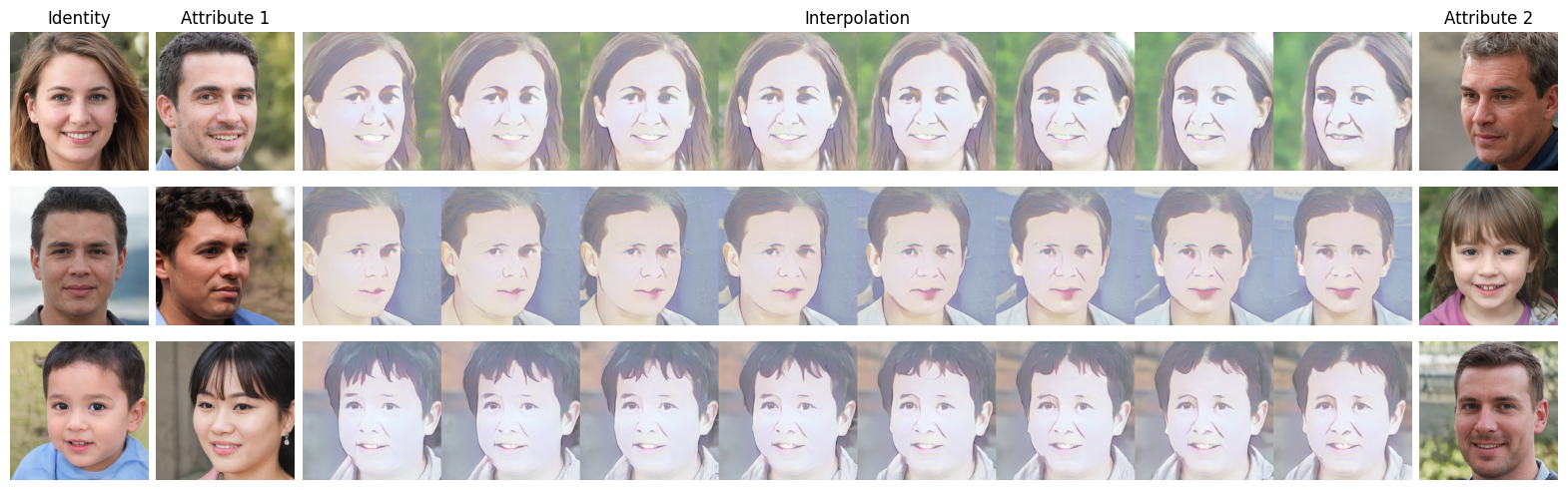}
    \caption{Results of pose/expression (first 3 rows) and identity (last 3 rows) interpolation. We sample eight vectors in the linearly interpolated space between the mapped latent codes $w_1$ and $w_2$ using $w = w_1 + (w_2 - w_1) \frac{k}{8}$ where $k \in \{1,2,...,8\}$, and pass it through our generator $C(G(w))$ to obtain the cartoonised images.}
    \label{interpolation}
\end{figure*}
The overall MLP loss is a weighted sum of the above losses:
\begin{equation}
\mathcal{L}_{total} = \lambda_{1}\mathcal{L}_{id} + \lambda_{2}\mathcal{L}_{lnd} + \lambda_{3}\mathcal{L}_{rec}   
\end{equation}

\section{Experiments}
\label{sec:experiments}

\textbf{Implementation : }We use StyleGAN pre-trained at 256x256 resolution in all our experiments, to easily compare with other methods. The ratio of training samples with $I_{id} = I_{p}$ and $I_{id} \neq I_{p}$ is an hyper-parameter that controls the weight for disentanglement and reconstruction. As suggested in \cite{Nitzan2020D}, we take $I_{id} = I_{p}$ every third iteration, and $I_{id} \neq I_{p}$ otherwise. $E_{lnd}$ is implemented using a pre-trained landmarks regression network \cite{landmarks}, trained to regress 68 facial keypoints.\\

\textbf{Hyper Parameters : }We use learning rate of $5e^{-5}$ when optimizing $\mathcal{L}_{total}$. Loss weights are set to $\lambda_{1} = 1, \lambda_{2} = 1, \lambda_{3} = 0.001$ and $\alpha = 0.84$.\\

\textbf{Dataset : }We create a dataset using StyleGAN in the way proposed in \cite{Richardson2020EncodingIS}. 
We sample 70,000 random Gaussian vectors and forward them through a pre-trained StyleGAN. In the forward process, the Gaussian noise is mapped into a latent vector , from which an image is generated, and we save the image and the vector as well. These vectors can be used for additional supervision to the MLP in the form of an adversarial loss but we found that it does not help our objective of cartoonisation (see Sec. \ref{abl}). \\

\textbf{Training : }Our training methodology is pretty straightforward. We begin by randomly selecting $I_{id}$ and $I_{p}$ images. When $I_{id} = I_{p}$ (i.e. both images are the same), the MLP learns to encode all the necessary information essential for achieving accurate reconstruction in the $W$ space, basically turning into a GAN inverter. When $I_{id} \neq I_{p}$ the MLP is taught to disentangle the intrinsic identity from other attributes present in the image. \\  

\subsection{Disentanglement of Identity and Pose}
As previously outlined, our approach leverages the inherent characteristics of the chosen generator, namely StyleGAN, and its associated latent space $W$. Previous works have extensively demonstrated that the latent space $W$ exhibits a high degree of controllability, allowing for seamless feature manipulation through latent code interpolation \cite{Shen2019InterpretingTL, Zhu2020InDomainGI}. Our proposed method discerns latent codes that are optimal for cartoonisation while also capturing the high-dimensional identity attribute and an intricate combination of expression and pose. \\
In Fig. \ref{interpolation}, we visually demonstrate the disentanglement of these high-level features through latent code interpolation, showcasing the robust nature of StyleGAN's latent space $W$ and its resilience to style distortion. In the first two rows of the figure, we keep the identity fixed and extract the attribute features (representing pose and expression) from two different images as labelled. We then proceed to obtain two corresponding latent codes $w_1$ and $w_2$ using our proposed method, and linearly interpolate between them. Similarly, for the last two rows we keep the attribute features fixed and interpolate between two different identities. \\
The generated cartoonised images exhibit a natural and smooth transition between the interpolated features while preserving the abstract style of cartoonisation. However, we observe minute variations in ideally constant facial attributes across the interpolated images which may lead to flickers while processing a sequence of frames.

\begin{figure}[ht!]\centering
\begin{tikzpicture}[zoomboxarray 2, 
                    zoomboxes 2 below]
    \node [image node 2] {\includegraphics[width=0.99\columnwidth]{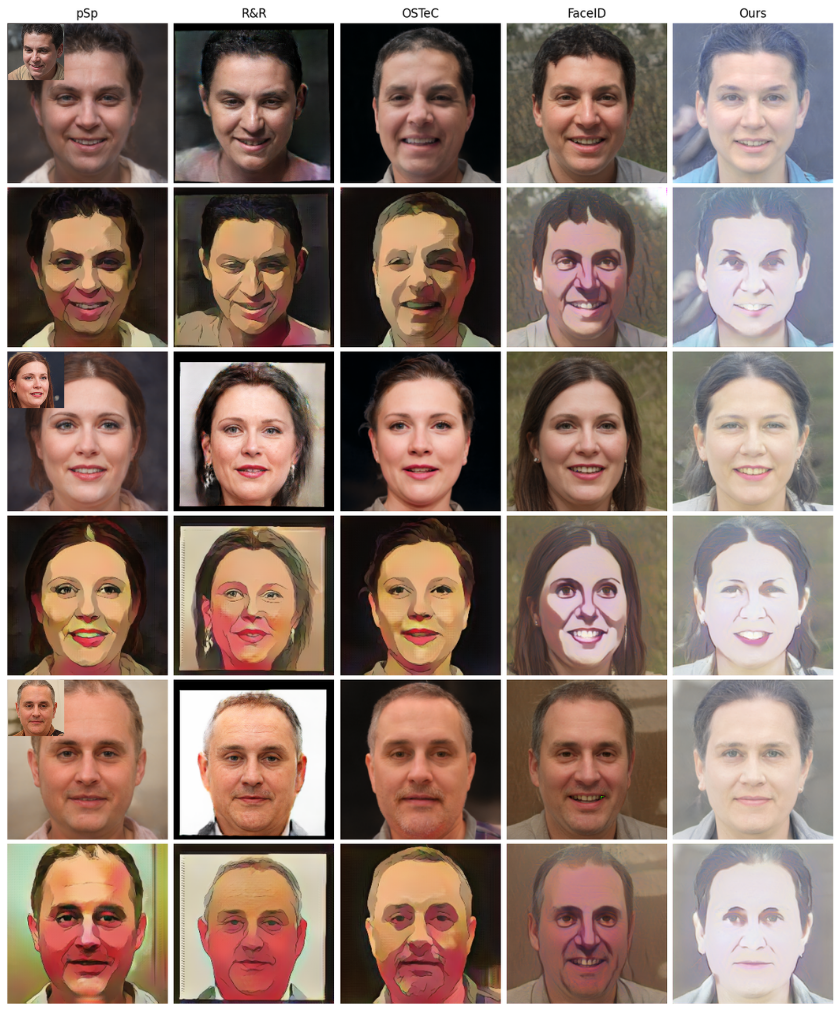}};
        \zoomboxtwo[magnification=6,color code 2=red]{0.07,0.73}
        \zoomboxtwo[magnification=6,color code 2=blue]{0.67,0.41}
        \zoomboxtwo[magnification=6,color code 2=green]{0.53,0.08}
        \zoomboxtwo[magnification=6,color code 2=red]{0.86,0.73}
        \zoomboxtwo[magnification=6,color code 2=blue]{0.86,0.41}
        \zoomboxtwo[magnification=6,color code 2=green]{0.93,0.08}
\end{tikzpicture}
\caption{(a) Comparison of our method with face frontalization models namely, pSp \cite{pix2style2pix}, R\&R \cite{RotateAndRender}, OSTeC \cite{ostec} and FaceID \cite{Nitzan2020D}, followed by successive cartoonisation \cite{Wang_2020_CVPR}. Notice how our intermittent frontal images are better suited for cartoonisation. (b) Other methods produce shading and texture artifacts especially around the eyes, cheeks and mouth regions. Our results illustrate better cartoonization characterized by well-defined edges, consistent and smooth shading, and plain textures.}
\label{frontal}
\end{figure}

\subsection{Loss In Identity Due To Cartoonisation}
\label{exp:identity}
Our approach possesses the capability to generate cartoonised versions of diverse identities across various poses. However, it's worth noting that, to our knowledge, there aren't existing models that explicitly cater to this specific task. In light of this, we conduct a comparative analysis by juxtaposing our results against those of other face frontalisation models \cite{pix2style2pix, RotateAndRender, ostec, Nitzan2020D} coupled with the White-Box Cartooniser \cite{Wang_2020_CVPR}, as shown in Fig. \ref{frontal}. This approach facilitates the cartoonisation of the outputs from these models, allowing us to gauge the impact on identity preservation as well as stylisation. \\
Our comparison revolves around quantifying the loss in identity as a consequence of this cartoonisation process. To calculate the loss in identity related information of $I_{x}$ after cartoonisation we subtract $\mathcal{L}_{id}(I_{x}, G(w))$ from $\mathcal{L}_{id}(I_{x}, G(C(w))$. Similarly for other frontalisation methods we calculate the loss in identity due to cartoonisation by subtracting $\mathcal{L}_{id}$ for each image after cartoonisation from $\mathcal{L}_{id}$ before. 
 Our findings listed in Table \ref{table:id}, clearly show that our model exhibits superior performance in terms of mitigating the increment in identity loss, after the cartoonisation step. This outcome shows the robustness of our approach and its effectiveness in maintaining identity-related information during the style transfer from realistic to cartoon aesthetics.\\

\renewcommand{\arraystretch}{1.25}
\begin{table}[ht!]
\centering
\begin{tabular}{@{}lccc@{}}
\hline
\multirow{2}{*}{Method} & \multicolumn{2}{c}{Average Identity Loss ($\overline{\mathcal{L}}_{id}$) ↓}                                                                                         & \multirow{2}{*}{Increment ↓} \\ \cline{2-3}
& \begin{tabular}[c]{@{}c@{}}Before \\ Cartoonisation\end{tabular} & \begin{tabular}[c]{@{}c@{}}After \\ Cartoonisation\end{tabular} &                             \\ \hline
pSp \cite{pix2style2pix}                     & 0.1548                                                           & \textbf{0.5092}                                                          & 0.3544                      \\
R\&R \cite{RotateAndRender}                    & 0.3010                                                           & 0.5797                                                          & 0.2787                      \\
OSTeC \cite{ostec}                   & \textbf{0.2326}                                                           & 0.5710                                                          & 0.3384                      \\
FaceID \cite{Nitzan2020D}                  & 0.3441                                                           & 0.7046                                                          & 0.3605                      \\
Ours         & 0.4903                            &                     0.6117                                                 & \textbf{0.1213}             \\ \hline
\end{tabular}
\caption{Quantitative evaluation of frontalisation and subsequent cartoonisation on StyleGAN generated data with different identities and poses.}
\label{table:id}
\end{table}


\subsection{Different $W$ Spaces}
\begin{figure}[h]
     \centering
     \begin{subfigure}[H]{0.45\textwidth}
         \centering
         \includegraphics[width=\textwidth]{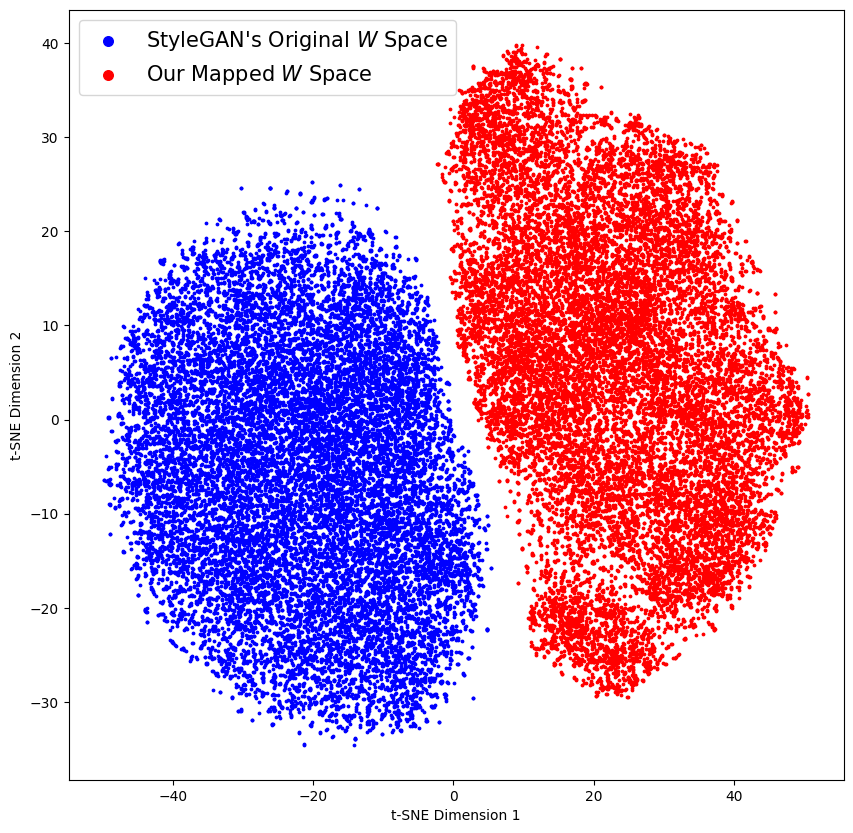}
         \caption{2-d t-SNE plot}
         \label{fig:y equals x}
     \end{subfigure}
     \hfill
     \begin{subfigure}[H]{0.45\textwidth}
         \centering
         \includegraphics[width=\textwidth]{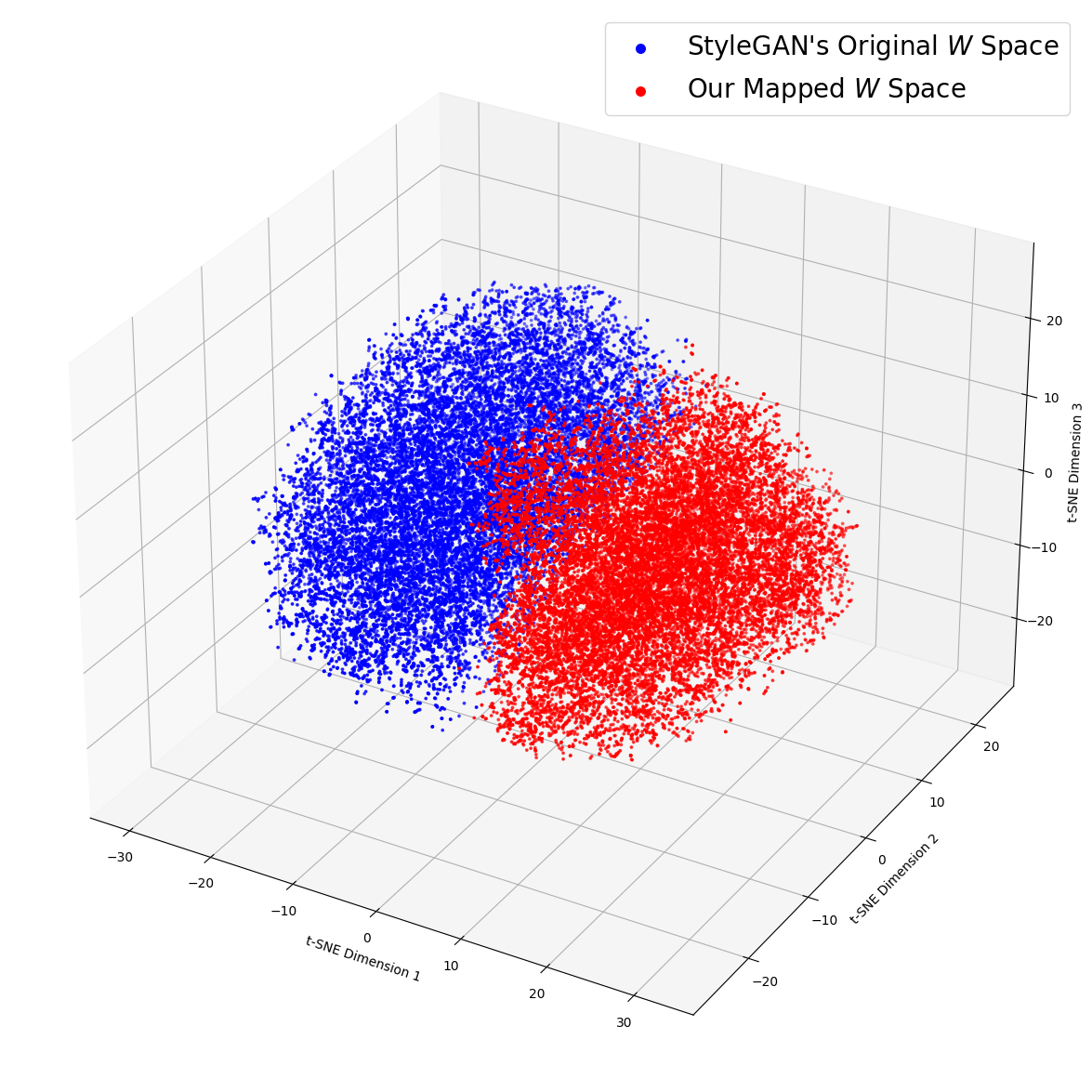}
         \caption{3-d t-SNE plot}
         \label{fig:three sin x}
     \end{subfigure}

        \caption{t-SNE visualizations of the latent vectors obtained from the two distributions : StyleGAN's original $W$ space and our mapped $W$ space}
        \label{fig:tsne}
\end{figure}
To validate our hypothesis that our encoder had indeed learned a novel distribution within the StyleGAN latent space, we devised a visualization strategy. We aimed to visually compare the original $w$ vectors with the vectors generated by our trained MLP. While we had previously demonstrated (as seen in Sec. \ref{exp:identity}) that the vectors produced by our MLP were better suited for maintaining identity after cartoonization, this time our focus was on confirming the distinctness of the two distributions – the MLP-generated vectors and the original StyleGAN $w$ vectors.

Our approach involved utilizing t-SNE visualizations \cite{tSNE} to depict the latent vectors derived from both distributions. Upon generating these t-SNE plots, discernible differences became apparent. The patterns and clusters formed by the StyleGAN-derived $w$ vectors exhibited clear separation from those formed by the vectors generated by our MLP. This visual divergence in distribution strongly supports the notion that the latent spaces of StyleGAN-generated images and the images produced by our encoder are not aligned as originally anticipated. This observation underscores the need for further exploration to understand the nuances contributing to this dissimilarity, potentially shedding light on the intricacies of our MLP's adaptation to the StyleGAN latent space.

\subsection{Ablations}
\label{abl}
\begin{figure*}[h]
     \centering
     \begin{subfigure}[H]{0.31\textwidth}
         \centering
         \includegraphics[width=\textwidth]{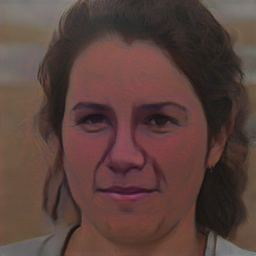}
         \includegraphics[width=\textwidth]{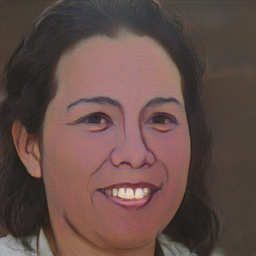}
         \caption{results without $\mathcal{L}_{rec}$}
         \label{fig:y equals x}
     \end{subfigure}
     \hfill
     \begin{subfigure}[H]{0.31\textwidth}
         \centering
         \includegraphics[width=\textwidth]{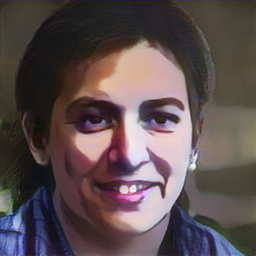}
         \includegraphics[width=\textwidth]{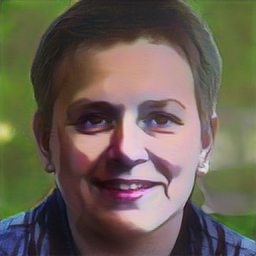}
         \caption{results with $\mathcal{L}_{adv}$}
         \label{fig:three sin x}
     \end{subfigure}
     \hfill
     \begin{subfigure}[H]{0.31\textwidth}
         \centering
         \includegraphics[width=\textwidth]{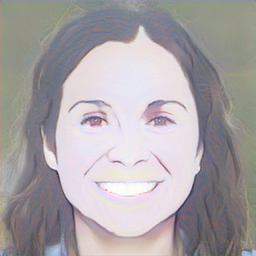}
         \includegraphics[width=\textwidth]{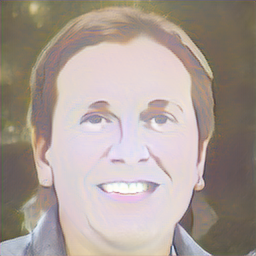}
         \caption{results without $\mathcal{L}_{lnd}$}
         \label{fig:five over x}
     \end{subfigure}
        \caption{Results of ablations, (a) Shows minor noisy details on the face, (b) Adding adversarial loss makes the output more realistic (c) Removing $\mathcal{L}_{lnd}$ makes all outputs have same pose. }
        \label{fig:three graphs}
\end{figure*}
To show the effectiveness of each component in our loss calculation method we perform experiments without individual components and report our results (see Fig. \ref{fig:three graphs} ) and observations. Ablating the recreational loss we found that smaller details that contribute to identity are lost and the overall the face looks messy (Fig. \ref{fig:y equals x}). Ablating the landmark loss causes the results to always have the same pose (Fig. \ref{fig:five over x}), meaning that enforcing landmark also helps us get better embeddings from our pose encoder that is being fine-tuned.\\
We also tried to use the randomly sampled $w$ vectors from the StyleGAN latent space as real samples to train a discriminator that tries to classify the real $w$ from the generated $w$. The output of the discriminator was used to calculate an adversarial loss term (as mentioned in \cite{Nitzan2020D}) which we added to the total loss. We found that the incorporation of the adversarial loss term adds to the inherent complexity of StyleGAN-generated images and the resulting outputs retained a substantial amount of intricate details that contribute to a heightened sense of realism in the images (as illustrated in Fig. \ref{fig:three sin x}). As a result, the final outputs after cartoonisation lacked the desired smooth textures that we set out to achieve. This intricate balance between retaining realism and simplifying textures led us to reevaluate the interplay between adversarial loss and texture preservation.

\section{Conclusion}
In this paper we proposed a method to achieve cartoonisation given an identity and a pose image by learning the separate representations for identity and pose in the $W$ latent space of StyleGAN in a semi supervised fashion. We used a readily available StyleGAN pretrained on FFHQ dataset with a white box cartoon generator which allowed us to bypass the fine-tuning process. We demonstrated that our encoder is better for encoding identity when cartoonisation is the objective by performing extensive experiments.\\
We leverage the generative abilities of the powerful pre-trained generator that we employed, but this involves not only adopting its strengths but also accepting its limitations, which occur because of the bias in training dataset. Due to the preprocessing methodology employed by StyleGAN which aligns the orientation of heads in images to eliminate roll angles, yaw rotations become closely linked with translation. Consequently, these characteristics are inherited by the images generated using StyleGAN, including our own generated outputs. Our method also leaves artefacts, usually associated with flickering in GANs, like hallucinating spectacles where there are none and changing hair structure with pose.\\  
The approach presented in this paper is agnostic to the type of style generator we use in the setup hence, this work can be extented to other styles like face sketch or line drawings. We have also used readily available standard models for calculating embeddings and we think there is scope of improvement in the quality of embeddings that we use to train the MLP. This work can also be extended to videos by introducing temporal losses and other flicker suppression mechanisms. StyleGAN is not capable of fully encompassing the entire spectrum of human head poses within its $W$ latent space. To address this, certain studies \cite{Abdal_2019_ICCV} have explored the concept of an augmented latent space referred to as $W+$. This expanded space enables the generator to produce images beyond human identities, encompassing a range of non-human subjects like animals and rooms, which is an exciting line of future work.     
{\small
\bibliographystyle{ieee_fullname}
\bibliography{PaperForReview}

\begin{thebibliography}{10}\itemsep=-1pt

\bibitem{Abdal_2019_ICCV}
Rameen Abdal, Yipeng Qin, and Peter Wonka.
\newblock Image2stylegan: How to embed images into the stylegan latent space?
\newblock In {\em Proceedings of the IEEE/CVF International Conference on
  Computer Vision (ICCV)}, October 2019.

\bibitem{cartoonGAN}
Yang Chen, Yu-Kun Lai, and Yong-Jin Liu.
\newblock Cartoongan: Generative adversarial networks for photo cartoonization.
\newblock In {\em Proceedings of the IEEE Conference on Computer Vision and
  Pattern Recognition (CVPR)}, June 2018.

\bibitem{EditingIS}
Edo Collins, Raja Bala, Bob Price, and Sabine S{\"u}sstrunk.
\newblock Editing in style: Uncovering the local semantics of gans.
\newblock {\em 2020 IEEE/CVF Conference on Computer Vision and Pattern
  Recognition (CVPR)}, pages 5770--5779, 2020.

\bibitem{ArcFaceAA}
Jiankang Deng, J. Guo, and Stefanos Zafeiriou.
\newblock Arcface: Additive angular margin loss for deep face recognition.
\newblock {\em 2019 IEEE/CVF Conference on Computer Vision and Pattern
  Recognition (CVPR)}, pages 4685--4694, 2018.

\bibitem{landmarks}
Zhenhua Feng, Josef Kittler, Muhammad Awais, P. Huber, and Xiaojun Wu.
\newblock Wing loss for robust facial landmark localisation with convolutional
  neural networks.
\newblock {\em 2018 IEEE/CVF Conference on Computer Vision and Pattern
  Recognition}, pages 2235--2245, 2017.

\bibitem{ostec}
Baris Gecer, Jiankang Deng, and Stefanos Zafeiriou.
\newblock Ostec: One-shot texture completion.
\newblock {\em CoRR}, abs/2012.15370, 2020.

\bibitem{goodfellow2014generative}
Ian~J Goodfellow, Jean Pouget-Abadie, Mehdi Mirza, Bing Xu, David Warde-Farley,
  Sherjil Ozair, Aaron Courville, and Yoshua Bengio.
\newblock Generative adversarial networks.
\newblock {\em arXiv preprint arXiv:1406.2661}, 2014.

\bibitem{MUNIT}
Xun Huang, Ming-Yu Liu, Serge~J. Belongie, and Jan Kautz.
\newblock Multimodal unsupervised image-to-image translation.
\newblock In {\em European Conference on Computer Vision}, 2018.

\bibitem{pix2pix2017}
Phillip Isola, Jun-Yan Zhu, Tinghui Zhou, and Alexei~A Efros.
\newblock Image-to-image translation with conditional adversarial networks.
\newblock {\em CVPR}, 2017.

\bibitem{StyleFusion}
Omer Kafri, Or Patashnik, Yuval Alaluf, and Daniel Cohen-Or.
\newblock Stylefusion: A generative model for disentangling spatial segments.
\newblock {\em ArXiv}, abs/2107.07437, 2021.

\bibitem{Karras_2019_CVPR}
Tero Karras, Samuli Laine, and Timo Aila.
\newblock A style-based generator architecture for generative adversarial
  networks.
\newblock In {\em Proceedings of the IEEE/CVF Conference on Computer Vision and
  Pattern Recognition (CVPR)}, June 2019.

\bibitem{UGATIT}
Junho Kim, Minjae Kim, Hyeonwoo Kang, and Kwanghee Lee.
\newblock U-gat-it: Unsupervised generative attentional networks with adaptive
  layer-instance normalization for image-to-image translation.
\newblock {\em ArXiv}, abs/1907.10830, 2019.

\bibitem{VGG}
Alex Krizhevsky, Ilya Sutskever, and Geoffrey Hinton.
\newblock Imagenet classification with deep convolutional neural networks.
\newblock {\em Neural Information Processing Systems}, 25, 01 2012.

\bibitem{inversion_new}
Zachary~Chase Lipton and Subarna Tripathi.
\newblock Precise recovery of latent vectors from generative adversarial
  networks.
\newblock {\em ArXiv}, abs/1702.04782, 2017.

\bibitem{UNIT}
Ming-Yu Liu, Thomas~M. Breuel, and Jan Kautz.
\newblock Unsupervised image-to-image translation networks.
\newblock {\em ArXiv}, abs/1703.00848, 2017.

\bibitem{FUNIT}
Ming-Yu Liu, Xun Huang, Arun Mallya, Tero Karras, Timo Aila, Jaakko Lehtinen,
  and Jan Kautz.
\newblock Few-shot unsupervised image-to-image translation.
\newblock {\em 2019 IEEE/CVF International Conference on Computer Vision
  (ICCV)}, pages 10550--10559, 2019.

\bibitem{Mirza2014ConditionalGA}
Mehdi Mirza and Simon Osindero.
\newblock Conditional generative adversarial nets.
\newblock {\em ArXiv}, abs/1411.1784, 2014.

\bibitem{Nitzan2020D}
Yotam Nitzan, Amit~H. Bermano, Yangyan Li, and Daniel Cohen-Or.
\newblock Disentangling in latent space by harnessing a pretrained generator.
\newblock {\em ArXiv}, abs/2005.07728, 2020.

\bibitem{Radford2015UnsupervisedRL}
Alec Radford, Luke Metz, and Soumith Chintala.
\newblock Unsupervised representation learning with deep convolutional
  generative adversarial networks.
\newblock {\em CoRR}, abs/1511.06434, 2015.

\bibitem{Richardson2020EncodingIS}
Elad Richardson, Yuval Alaluf, Or Patashnik, Yotam Nitzan, Yaniv Azar, Stav
  Shapiro, and Daniel Cohen-Or.
\newblock Encoding in style: a stylegan encoder for image-to-image translation.
\newblock {\em 2021 IEEE/CVF Conference on Computer Vision and Pattern
  Recognition (CVPR)}, pages 2287--2296, 2020.

\bibitem{pix2style2pix}
Elad Richardson, Yuval Alaluf, Or Patashnik, Yotam Nitzan, Yaniv Azar, Stav
  Shapiro, and Daniel Cohen{-}Or.
\newblock Encoding in style: a stylegan encoder for image-to-image translation.
\newblock {\em CoRR}, abs/2008.00951, 2020.

\bibitem{Shen2019InterpretingTL}
Yujun Shen, Jinjin Gu, Xiaoou Tang, and Bolei Zhou.
\newblock Interpreting the latent space of gans for semantic face editing.
\newblock {\em 2020 IEEE/CVF Conference on Computer Vision and Pattern
  Recognition (CVPR)}, pages 9240--9249, 2019.

\bibitem{FaceIDGANLA}
Yujun Shen, Ping Luo, Junjie Yan, Xiaogang Wang, and Xiaoou Tang.
\newblock Faceid-gan: Learning a symmetry three-player gan for
  identity-preserving face synthesis.
\newblock {\em 2018 IEEE/CVF Conference on Computer Vision and Pattern
  Recognition}, pages 821--830, 2018.

\bibitem{tSNE}
Laurens van~der Maaten and Geoffrey~E. Hinton.
\newblock Visualizing data using t-sne.
\newblock {\em Journal of Machine Learning Research}, 9:2579--2605, 2008.

\bibitem{Wang_2020_CVPR}
Xinrui Wang and Jinze Yu.
\newblock Learning to cartoonize using white-box cartoon representations.
\newblock In {\em IEEE/CVF Conference on Computer Vision and Pattern
  Recognition (CVPR)}, June 2020.

\bibitem{DualGAN}
Zili Yi, Hao Zhang, Ping Tan, and Minglun Gong.
\newblock Dualgan: Unsupervised dual learning for image-to-image translation.
\newblock {\em 2017 IEEE International Conference on Computer Vision (ICCV)},
  pages 2868--2876, 2017.

\bibitem{ssim}
Hang Zhao, Orazio Gallo, Iuri Frosio, and Jan Kautz.
\newblock Loss functions for image restoration with neural networks.
\newblock {\em IEEE Transactions on Computational Imaging}, 3:47--57, 2017.

\bibitem{RotateAndRender}
Hang Zhou, Jihao Liu, Ziwei Liu, Yu Liu, and Xiaogang Wang.
\newblock Rotate-and-render: Unsupervised photorealistic face rotation from
  single-view images.
\newblock {\em CoRR}, abs/2003.08124, 2020.

\bibitem{Zhu2020InDomainGI}
Jiapeng Zhu, Yujun Shen, Deli Zhao, and Bolei Zhou.
\newblock In-domain {GAN} inversion for real image editing.
\newblock {\em CoRR}, abs/2004.00049, 2020.

\bibitem{inversion}
Jun-Yan Zhu, Philipp Kr{\"a}henb{\"u}hl, Eli Shechtman, and Alexei~A. Efros.
\newblock Generative visual manipulation on the natural image manifold.
\newblock In {\em European Conference on Computer Vision}, 2016.

\bibitem{cyclegan}
Jun-Yan Zhu, Taesung Park, Phillip Isola, and Alexei~A. Efros.
\newblock Unpaired image-to-image translation using cycle-consistent
  adversarial networks.
\newblock {\em 2017 IEEE International Conference on Computer Vision (ICCV)},
  pages 2242--2251, 2017.

\end{thebibliography}
}

\end{document}


\title{Face Cartoonisation For Various Poses Using StyleGAN\\
Supplementary Material}

\author{First Author\\
Institution1\\
Institution1 address\\
{\tt\small firstauthor@i1.org}
\and
Second Author\\
Institution2\\
First line of institution2 address\\
{\tt\small secondauthor@i2.org}
}

\twocolumn[{
\maketitle
\begin{center}
\centering
\captionsetup{type=figure}\addtocounter{figure}{-1}

     \centering
     \begin{subfigure}[H]{0.23\textwidth}
         \centering
         \includegraphics[width=\textwidth]{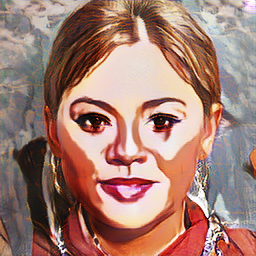}
         \caption{Result after 10k iterations}
         \label{fig:ld loss}
     \end{subfigure}
     \hfill
     \begin{subfigure}[H]{0.23\textwidth}
         \centering
         \includegraphics[width=\textwidth]{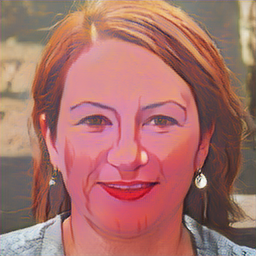}
         \caption{Result after 20k iterations}
         \label{fig:lnd loss}
     \end{subfigure}
     \hfill
     \begin{subfigure}[H]{0.23\textwidth}
         \centering
         \includegraphics[width=\textwidth]{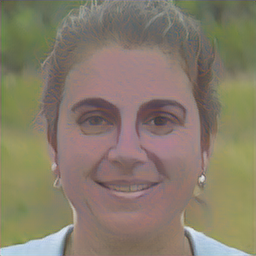}
         \caption{Result after 30k iterations}
         \label{fig:recgraph}
     \end{subfigure}
     \hfill
     \begin{subfigure}[H]{0.23\textwidth}
         \centering
         \includegraphics[width=\textwidth]{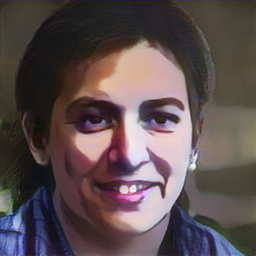}
         \caption{Result after 40k iterations}
         \label{fig:recgraph}
     \end{subfigure}

\captionof{figure}{This figure shows intermittent results when training with adversarial loss. The fluctuating results in this figure illustrate the instability induced by the loss function. The highly varied outcomes indicate that the adversarial loss, which aims to balance realism and accuracy, can lead to erratic model behavior. This instability is reflected in the inconsistent quality of the generated outputs across different training iterations, highlighting the challenge of achieving a stable and consistent performance when using this loss function.}
\label{illus_gan}
\end{center}
}]
\tableofcontents

\section{Effect of Adversarial Loss}
\begin{figure}[h]
     \centering
     \begin{subfigure}[H]{0.45\textwidth}
         \centering
         \includegraphics[width=\textwidth]{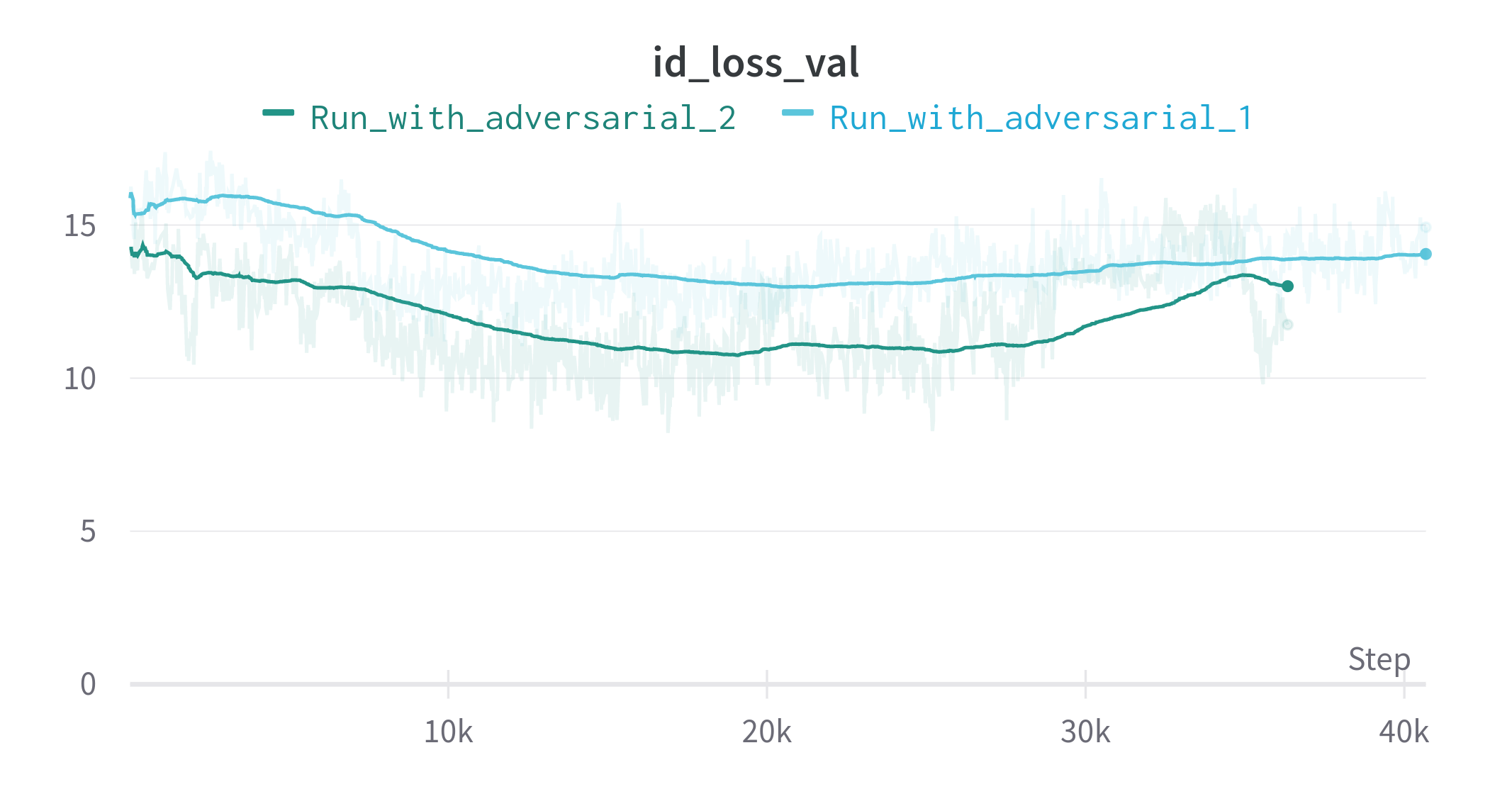}
        
         \label{fig:ld loss}
     \end{subfigure}
     \hfill
     \begin{subfigure}[H]{0.45\textwidth}
         \centering
         \includegraphics[width=\textwidth]{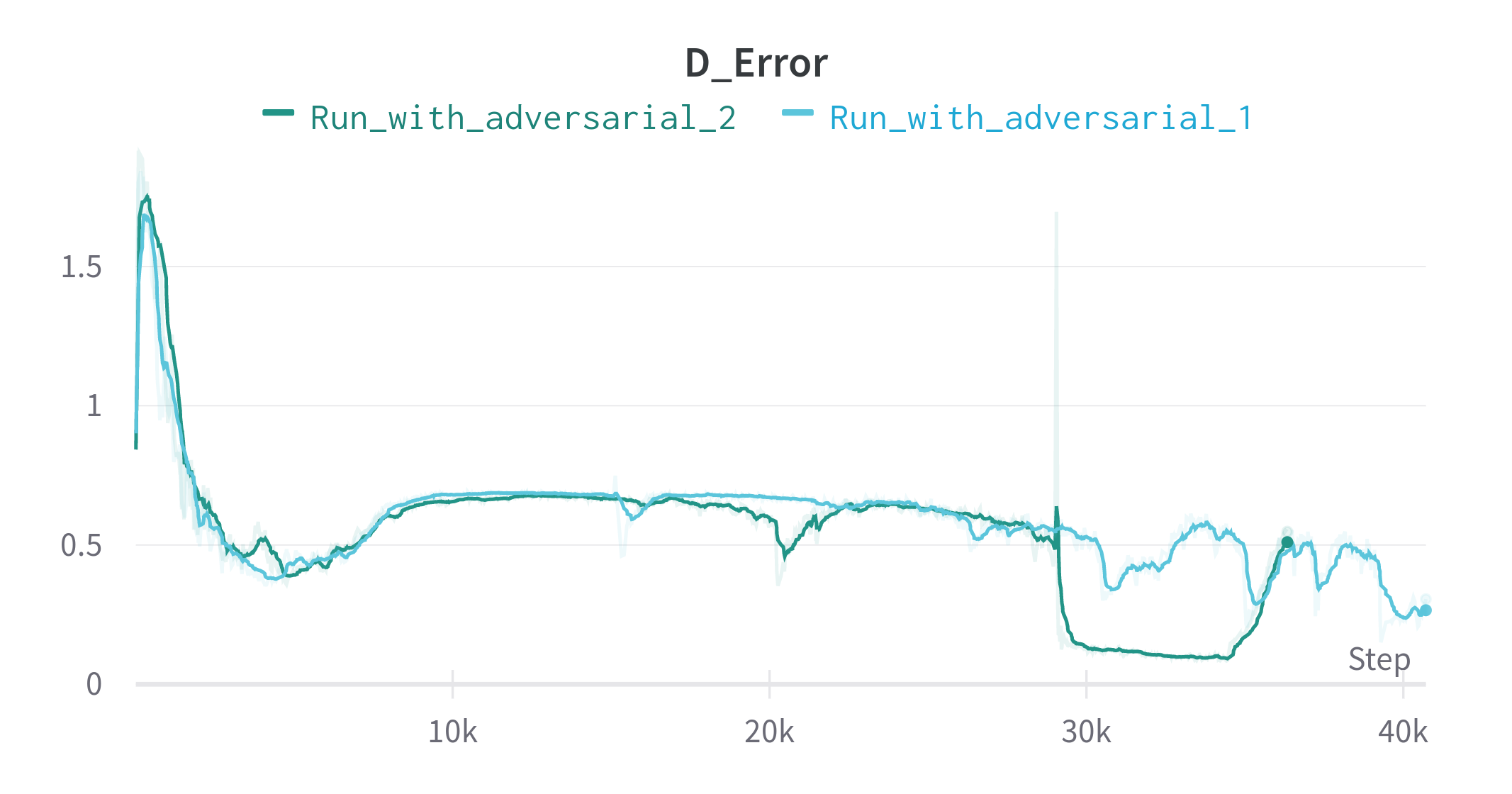}
         
     \end{subfigure}
     \hfill
     \begin{subfigure}[H]{0.45\textwidth}
         \centering
         \includegraphics[width=\textwidth]{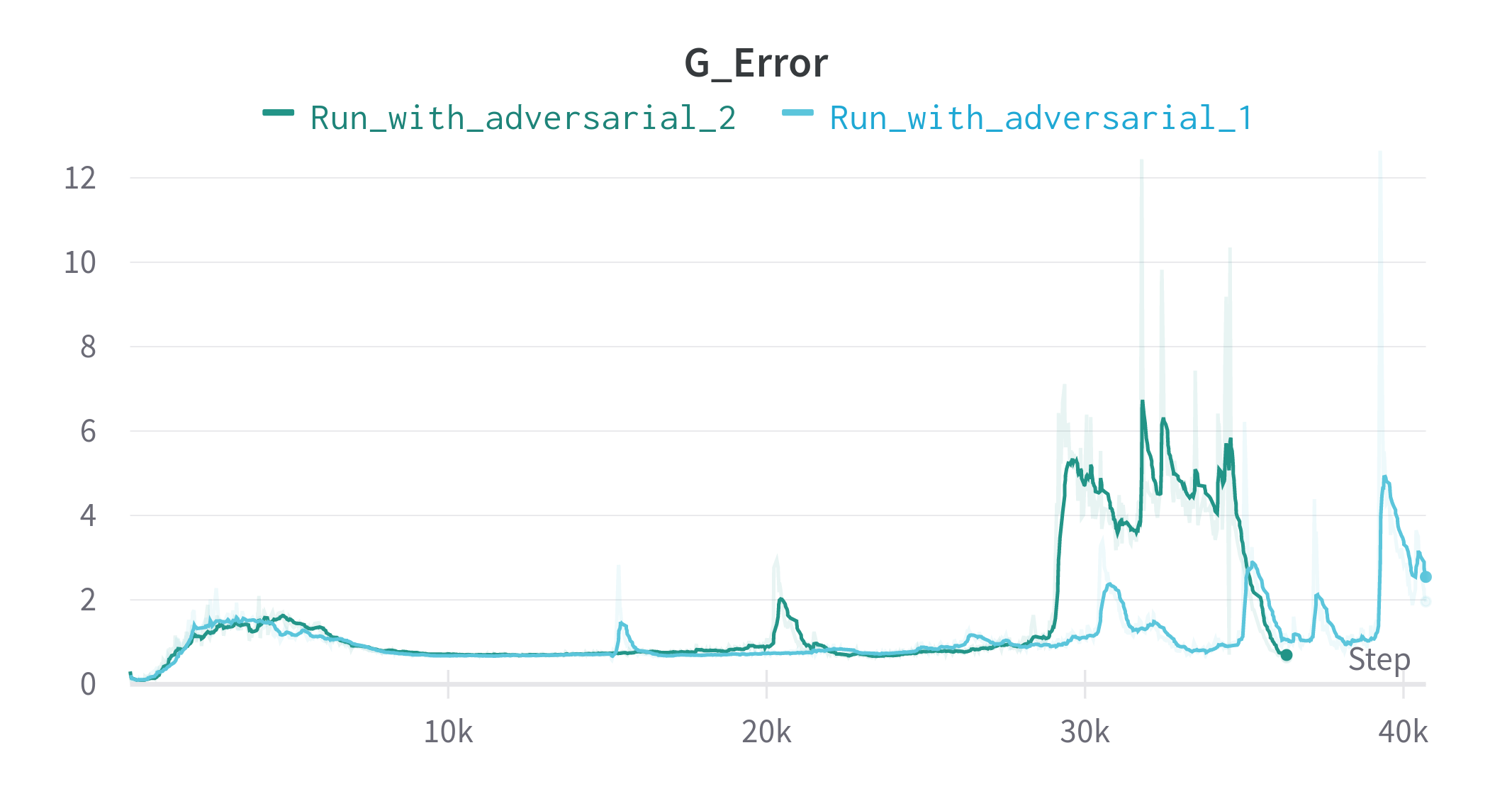}
         
         \label{fig:recgraph}
     \end{subfigure}

     \hfill
     \begin{subfigure}[H]{0.45\textwidth}
         \centering
         \includegraphics[width=\textwidth]{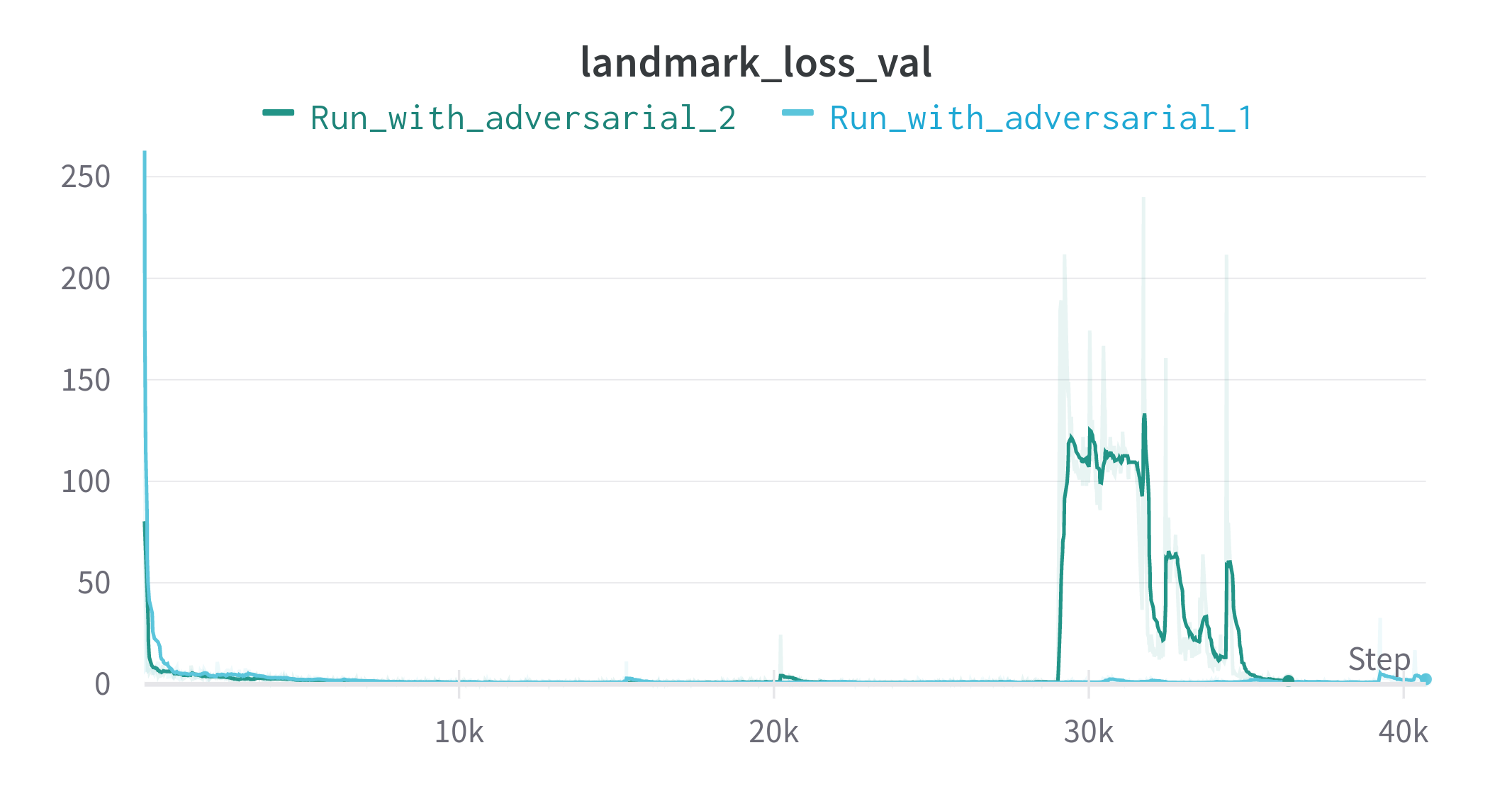}
         
         \label{fig:recgraph}
     \end{subfigure}

        \caption{As the training iterations progress, the model's performance oscillates, leading to irregular peaks and troughs in these loss values, confirming adversarial loss instability}
        \label{fig_loss}
\end{figure}
Making use of the sampled $w$ vectors as supervison for the discriminator network, we tried using the adversarial loss to confine the generator to a realistic latent space. We used the following as losses : \\
$$\begin{gathered}\mathcal{L}_{adv}^{D}=-\underset{w\sim\mathcal{W}}{\mathbb{E}}[logD_{\mathcal{W}}(w)]-\underset{z}{\mathbb{E}}[log(1-D_{\mathcal{W}}(M(z)))]+\\
\frac{\gamma}{2}\underset{w\sim\mathcal{W}}{\mathbb{E}}\left[\left\lVert\nabla_{w}D_{\mathcal{W}}(w)\right\rVert_{2}^{2}\right]\end{gathered}$$
$$\mathcal{L}_{adv}^{G}=-\underset{z}{\mathbb{E}}[logD_{\mathcal{W}}(M(z))]$$
As evident from the figure \ref{illus_gan} depicting intermittent training results, the adversarial loss introduces a significant level of instability. The fluctuating outcomes vividly illustrate the challenge of striking a delicate balance between realism and accuracy. This instability isn't confined to the generated images alone; it permeates throughout the model's behavior (figure \ref{fig_loss}), leading to erratic performance. Moreover, the accompanying graphs for generator error and discriminator error reaffirm this behavior, providing quantitative evidence of the adversarial loss's influence. Even though we explained why we did not use this loss in the paper, this is another reason why we decided to not  include adversarial loss.

\section{Training Details}
\begin{figure}[h]
     \centering
     \begin{subfigure}[H]{0.45\textwidth}
         \centering
         \includegraphics[width=\textwidth]{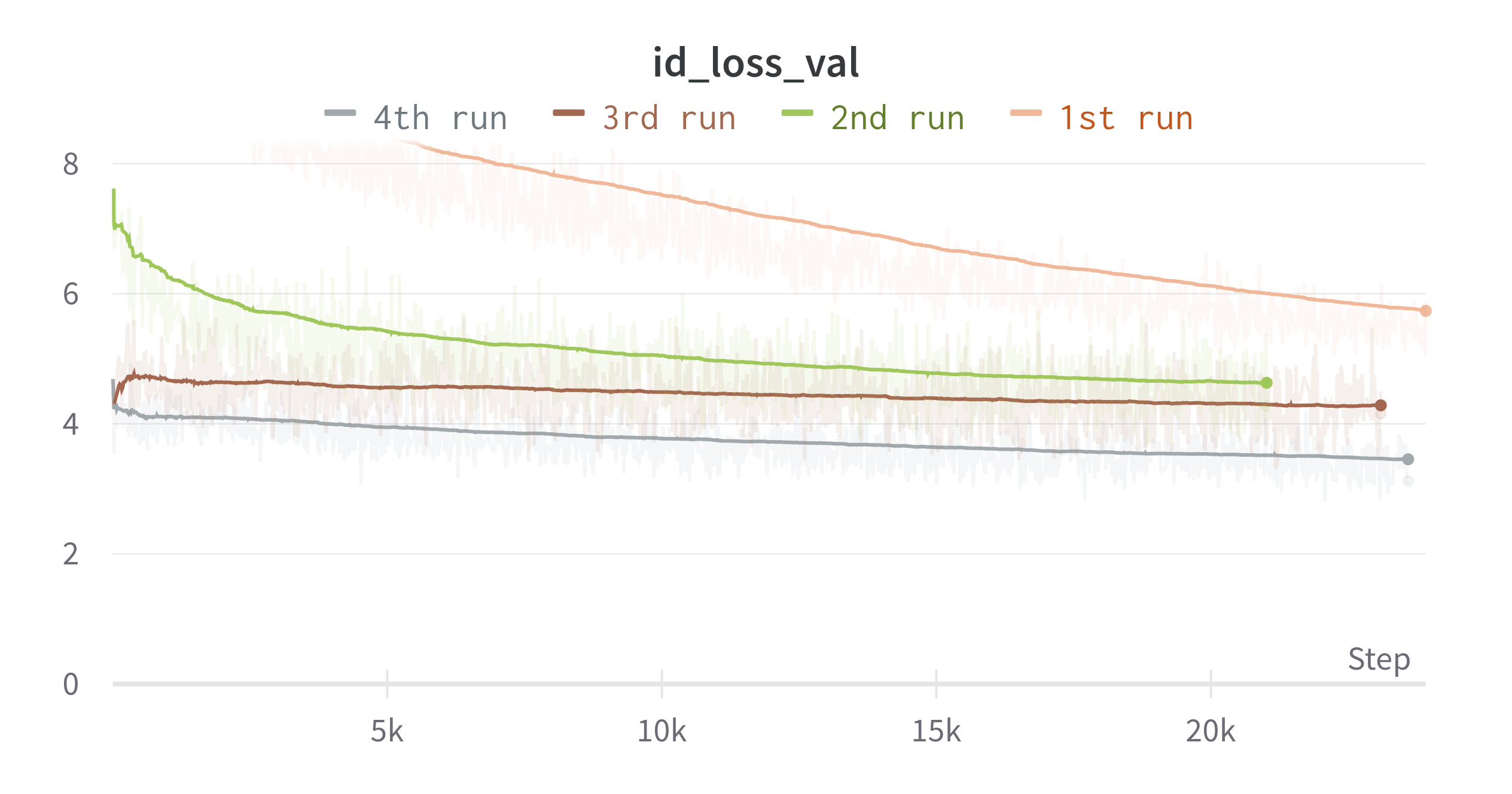}
         \caption{Plot of identity loss}
         \label{fig:ld loss}
     \end{subfigure}
     \hfill
     \begin{subfigure}[H]{0.45\textwidth}
         \centering
         \includegraphics[width=\textwidth]{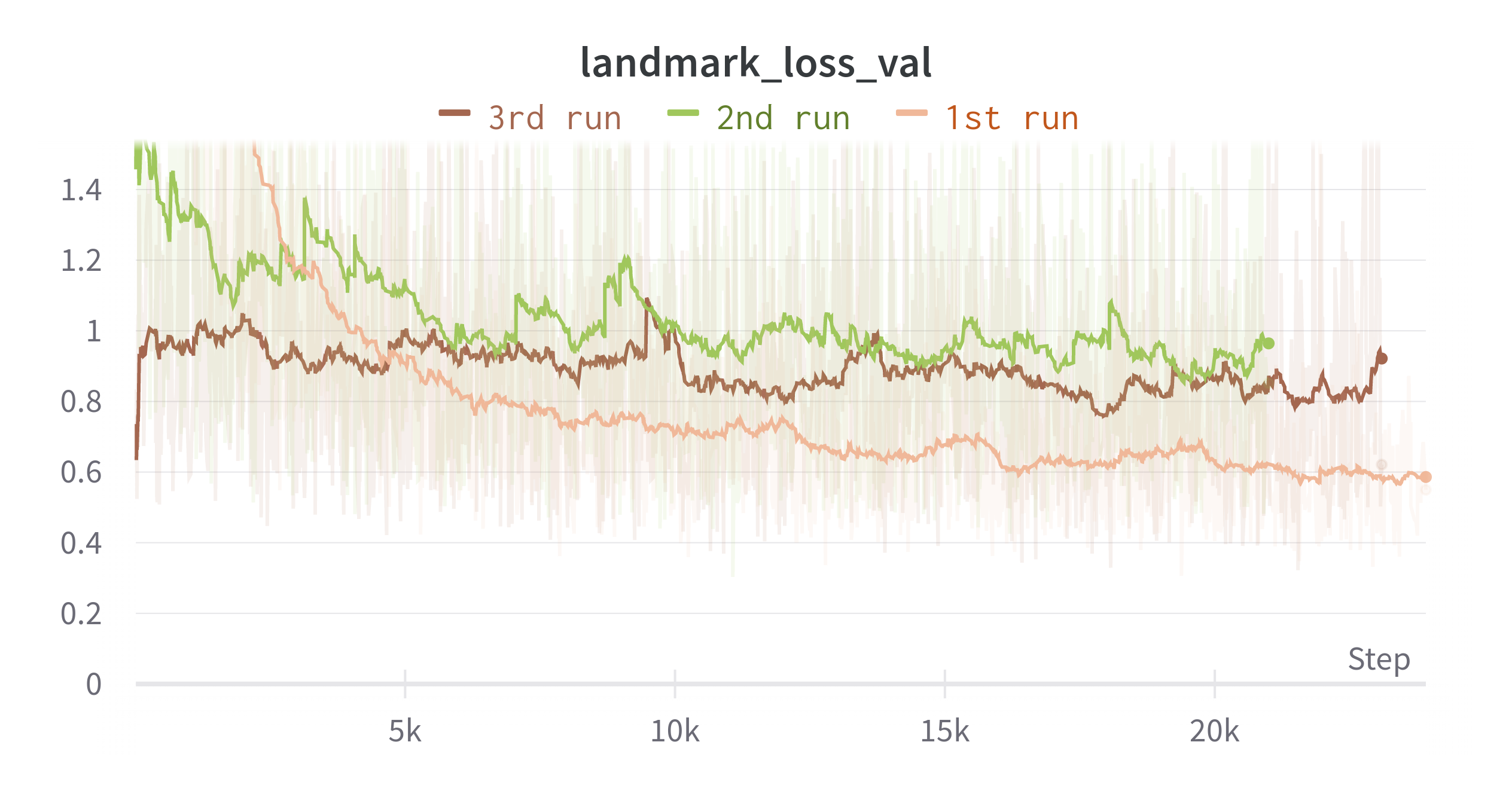}
         \caption{Plot of landmark loss}
         \label{fig:lnd loss}
     \end{subfigure}
     \hfill
     \begin{subfigure}[H]{0.45\textwidth}
         \centering
         \includegraphics[width=\textwidth]{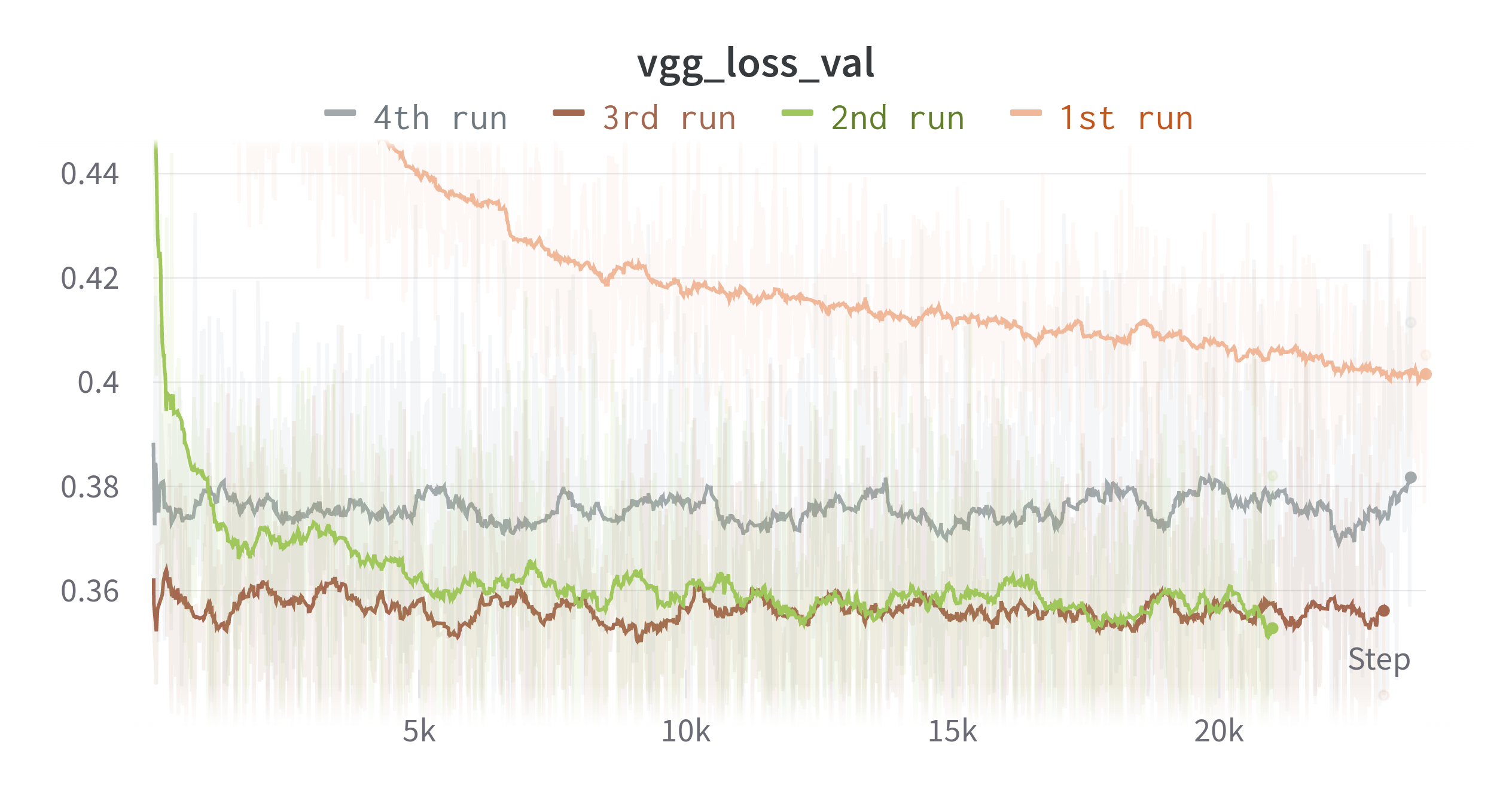}
         \caption{Plot of reconstruction loss}
         \label{fig:recgraph}
     \end{subfigure}

        \caption{Graphs of losses during our training process.}
        \label{figtraui}
\end{figure}
We train our model, using a single NVIDIA GeForce GTX 1080 Ti, in 4 consecutive runs of 6 hours each as shown in the figure \ref{figtraui}. We decided to not include $\mathcal{L}_{lnd}$ in the last run as it had already converged and this also helped $\mathcal{L}_{id}$ converge faster in the last 6 hours.

\begin{figure*}[ht]\centering
\includegraphics[width=0.8\textwidth]{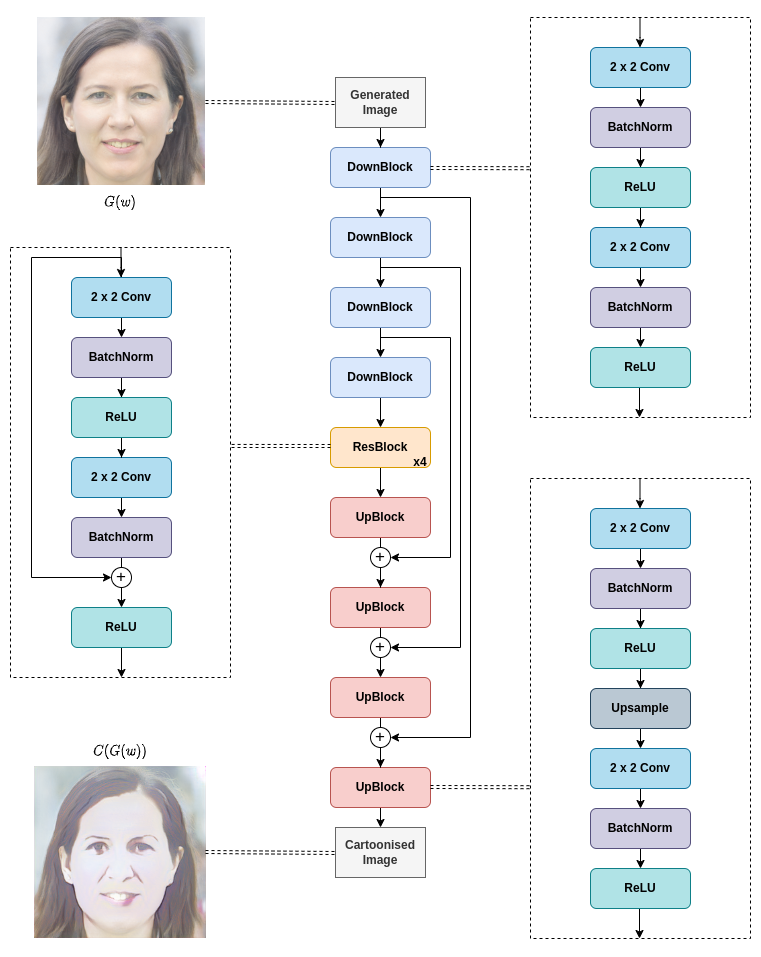}
\caption{The detailed architecture of our cartooniser $C$ (middle). The architecture of a residual block (left) and downsampling and upsampling blocks (right) used in $C$.}
\label{cartooniser_arch}
\end{figure*}

\section{Cartooniser GAN Framework}
As shown experimentally in \cite{DualStyleGAN}, the fine-resolution layers of StyleGAN capture the low-level colour details whereas coarse-resolution layers model the high level shape and structural style of the image. We utilise this characteristic of StyleGAN to choose a pretrained Cartooniser that has already learned the features relevant to color transfer in the finer layers to better guide the optimization objective of cartoonisation. In Fig. \ref{cartooniser_arch}, we present a detailed architecture of our cartooniser $C$, an essential component of our proposed generator. The cartooniser $C$ takes a StyleGAN generated image as input and outputs a caroonised version of the image. Improving on the fully convolutional U-Net-like \cite{u-net} architecture of the White Box Cartooniser \cite{Wang_2020_CVPR}, $C$ adopts a more extensive downsampling and upsampling framework. We use four downsampling blocks each consisting of two repeated applications of a 2x2 convolution, followed by 2D batch normalisation and activation using a ReLU unit. To ensure downsampling in each block, the first convolution uses a stride of 2 while the second convolution uses a stride of 1. Similarly, we use four upsampling blocks whose structure is identical to that of a downsampling block with the addition of an upsampling operation by a scale factor of 2 between two units of Conv-BatchNorm-ReLU layers where each convolution has a stride of 1. The bottleneck layer consists of four residual blocks similar to the original architecture.

\section{Limitations}

\begin{figure}[ht]\centering
\includegraphics[width=\columnwidth]{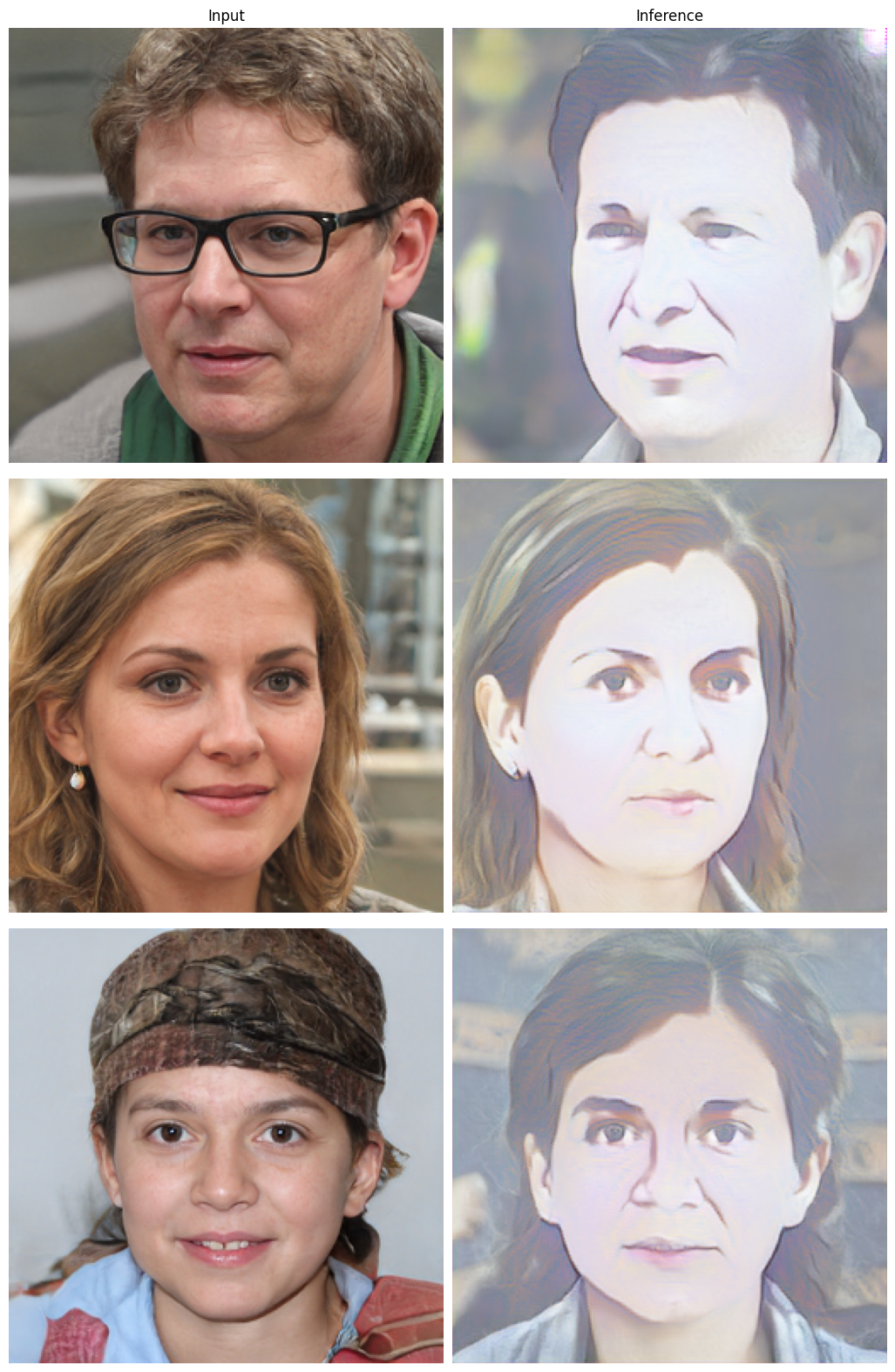}
\caption{Inference results to demonstrate accessory artifacts like spectacles, earrings and headgear.}
\label{lim1}
\end{figure}

\begin{figure}[ht]\centering
\includegraphics[width=\columnwidth]{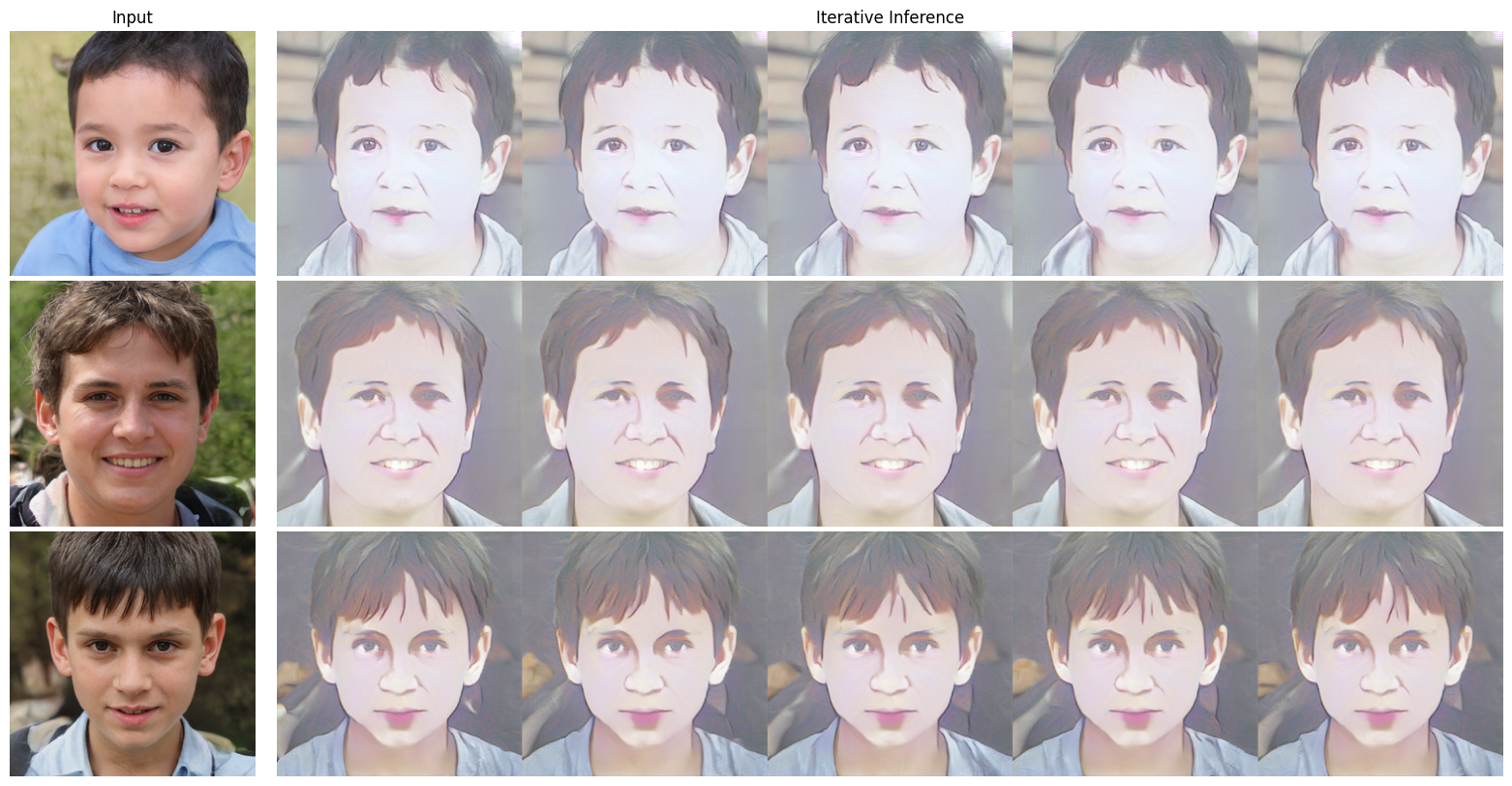}
\caption{We run inference multiple times on the same input image to show minor variation in details in the identity which may cause flickering.}
\label{lim2}
\end{figure}

\begin{figure}[h]
     \centering
     \begin{subfigure}[H]{\columnwidth}
         \centering
         \includegraphics[width=\textwidth]{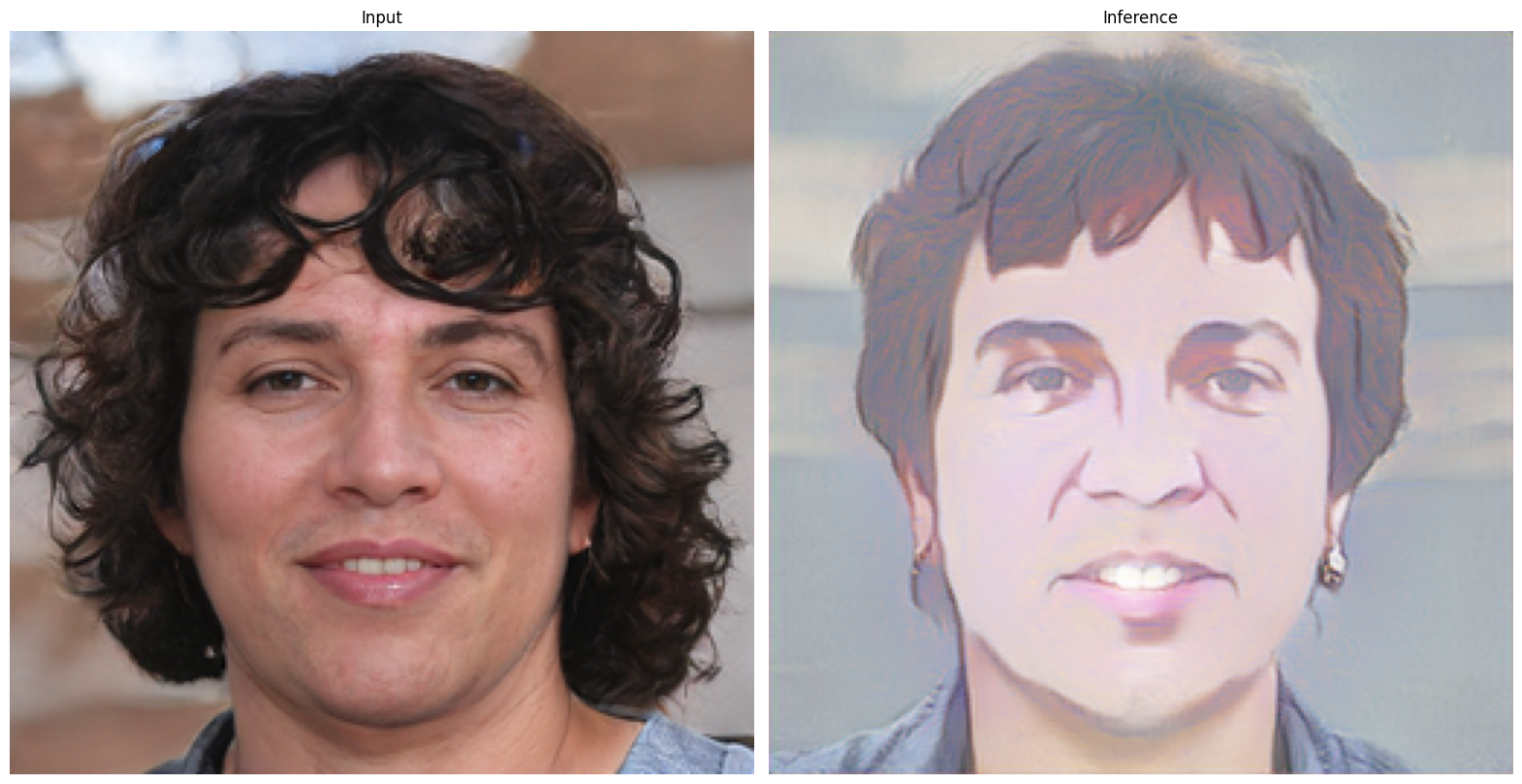}
         \caption{}
         \label{fig:lim3_a}
     \end{subfigure}
     \hfill
     \begin{subfigure}[H]{0.45\textwidth}
         \centering
         \includegraphics[width=\textwidth]{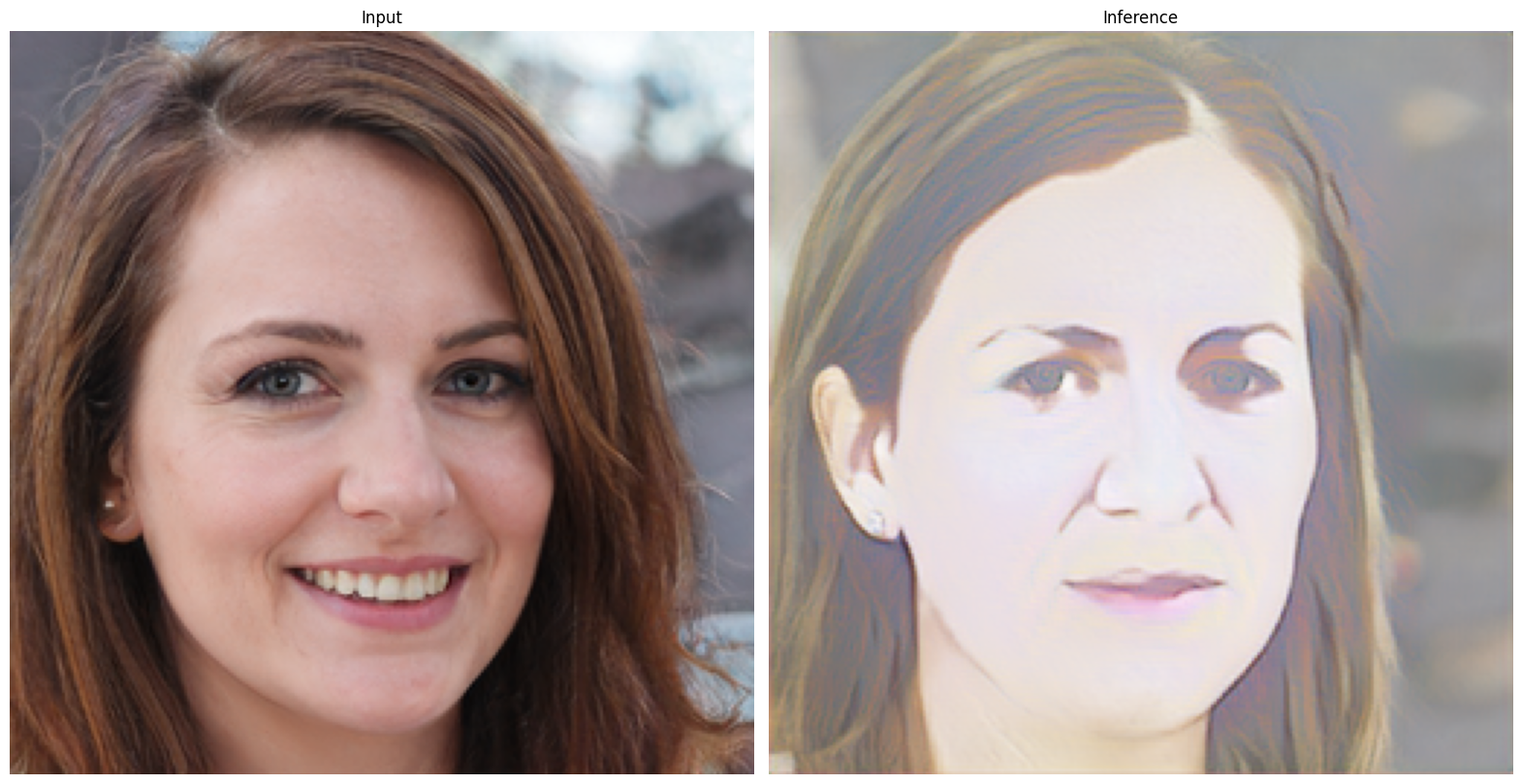}
         \caption{}
         \label{fig:lim3_b}
     \end{subfigure}
     \hfill
     \begin{subfigure}[H]{0.45\textwidth}
         \centering
         \includegraphics[width=\textwidth]{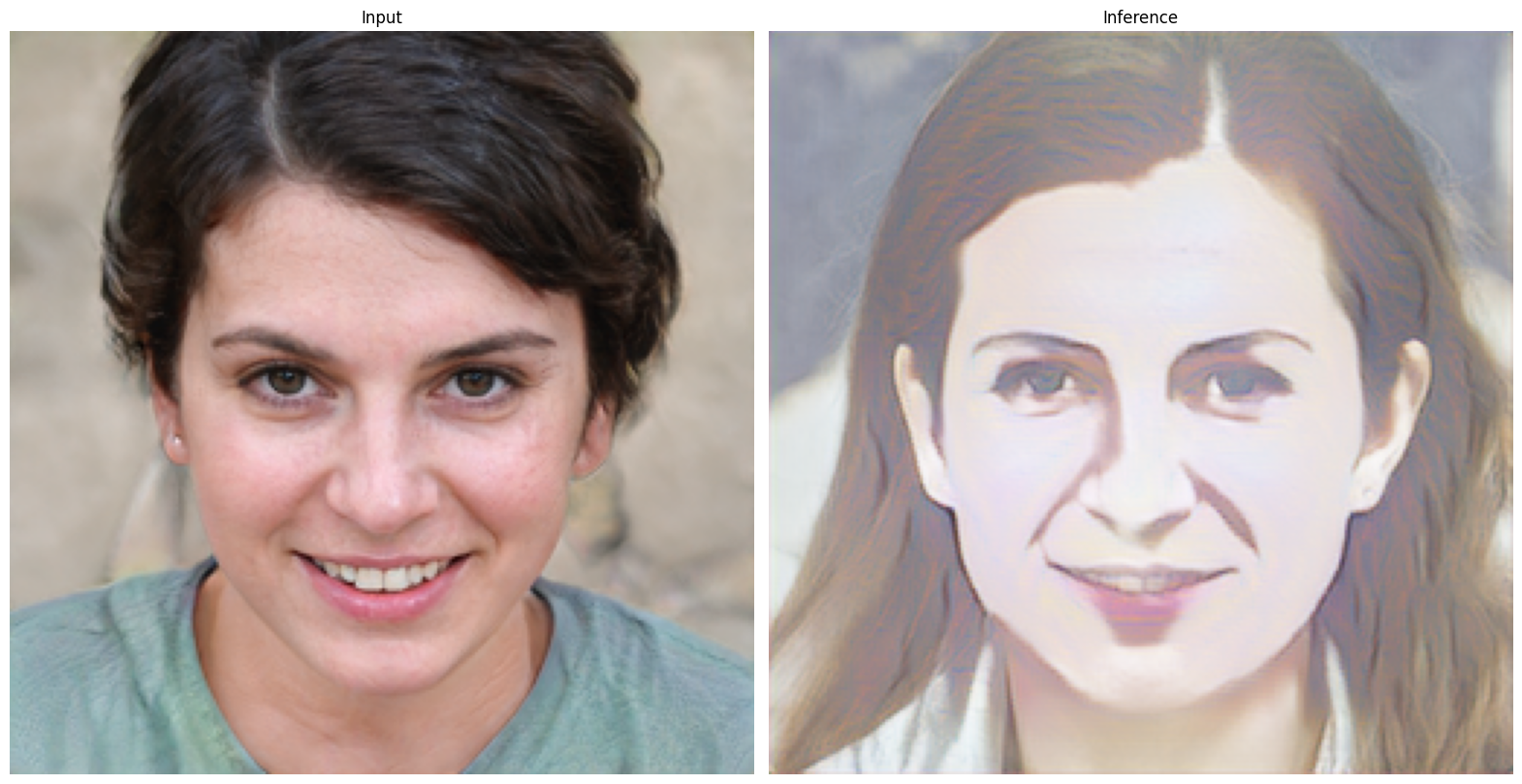}
         \caption{}
         \label{fig:lim3_c}
     \end{subfigure}

        \caption{More inference results to illustrate artifacts produced due to bias in dataset.}
        \label{fig:lim3}
\end{figure}

Our method inherits the generative abilities of the used pretrained generator, alongside which we also inherit any limitations that the generative modules might have, especially ones imposed by their training datasets. Our MLP is incapable of capturing facial accessories such as spectacles, earrings and headgear as demonstrated in Fig. \ref{lim1}. A common discrepancy that has been observed is the generation of inconsistent hairstyles as outlined in Fig. \ref{fig:lim3}. The colour and structure of hair is not transferred very well by our encoder hence our results have very little variation in hair colour and misrepresented hair structure (see Fig. \ref{fig:lim3}). More specifically, as seen in Fig. \ref{fig:lim3_b} and \ref{fig:lim3_c}, the orientation of hairstyle in the input images (left to right) is completely swapped (right to left). Furthermore, we also observe that our pose encoder does not capture the expression of a person quite accurately resulting in inconsistencies (See Fig. \ref{fig:lim3_b}). All these artifacts reflect apparent biases present in the training dataset.\\
Another striking limitation is the presence of minute inconsistencies during the generation process. To demonstrate this, we run multiple iterations of inference on the same input image. As observed in Fig. \ref{lim2}, there are granular discrepancies in the hair structure leading to temporal inconsistencies and flickering when a sequence of frames are processed.

{\small
\bibliographystyle{ieee_fullname}
\bibliography{egbib}
}